%% file: 0-main.tex
\documentclass{article}

\usepackage[preprint, nonatbib]{neurips_2023}

\usepackage[utf8]{inputenc} % allow utf-8 input
\usepackage[T1]{fontenc}    % use 8-bit T1 fonts
\usepackage{hyperref}       % hyperlinks
\usepackage{url}            % simple URL typesetting
\usepackage{booktabs}       % professional-quality tables
\usepackage{amsfonts}       % blackboard math symbols
\usepackage{nicefrac}       % compact symbols for 1/2, etc.
\usepackage{microtype}      % microtypography
\usepackage{xcolor}         % colors
\usepackage{amsthm}
\usepackage{graphicx}      % graphics
\usepackage{subcaption}
\usepackage[nonatbib]{neurips_2023}
\usepackage{bbm} %Mo
\usepackage{amsmath}
\DeclareMathOperator*{\argmax}{arg\,max}
\DeclareMathOperator*{\argmin}{arg\,min}
\usepackage{mathtools} %Mo
\usepackage{algorithm}      %Mo 
\usepackage[noend]{algpseudocode}

\usepackage{xcolor}

\newcommand{\algnamenospace}{BanditMIPS}
\newcommand{\algname}{BanditMIPS }

\include{_1-defs}

\renewcommand{\paragraph}[1]{\noindent {\bf #1}}
\renewcommand\thefootnote{*}

\title{Faster Maximum Inner Product Search in High Dimensions}

\author{%
  Mo Tiwari$^1$  \\
  \And
  Ryan Kang$^{1,*}$  \\
  \And
  Je-yong Lee$^{2,*}$  \\
  \And
  Donghyun Lee$^{3,*}$  \\
  \And
  Chris Piech$^1$ \\
  \And
  Sebastian Thrun$^1$ \\
  \And
  Ilan Shomorony$^{3, \#}$ \\
  \And
  Martin Jinye Zhang$^{4, \#}$ \\
}

\begin{document}

\maketitle

\begin{abstract}
Maximum Inner Product Search (MIPS) is a ubiquitous task in machine learning applications such as recommendation systems. 
Given a query vector and $n$ atom vectors in $d$-dimensional space, the goal of MIPS is to find the atom that has the highest inner product with the query vector. 
Existing MIPS algorithms scale at least as $O(\sqrt{d})$, which becomes computationally prohibitive in high-dimensional settings.
% Maximum Inner Product Search (MIPS) is a popular problem in the machine learning literature.
% In high-dimensional settings, however, MIPS queries can become computationally expensive as existing solutions do not scale well with dataset dimensionality.
In this work, we present \algnamenospace, a novel randomized MIPS algorithm whose complexity is independent of $d$. 
\algname estimates the inner product for each atom by subsampling coordinates and adaptively evaluates more coordinates for more promising atoms. The specific adaptive sampling strategy is motivated by multi-armed bandits. 
% We present a state-of-the-art algorithm for MIPS in high dimensions, dubbed \algnamenospace.
% \algname is a randomized algorithm that borrows techniques from multi-armed bandits to reduce the MIPS problem to a best-arm identification problem.
% \algname is more adaptive to the underlying data distribution than previous approaches because its complexity is governed by the average coordinate-wise gap $\Delta$ between the highest and second-highest inner products. When $\Delta$ does not depend on the dataset dimensionality $d$,
% \algname lowers the scaling complexity in $d$ of state-of-the-art algorithms from $O(\sqrt{d})$ to $O(1)$.
We provide theoretical guarantees that \algname returns the correct answer with high probability, while improving the complexity in $d$ from $O(\sqrt{d})$ to $O(1)$. 
%In other words, the runtime of \algname has \textit{no} explicit dependence on the data dimensionality; instead, its runtime is governed by the hardness of the underlying data distribution.
We also perform experiments on four synthetic and real-world datasets and demonstrate that \algname outperforms prior state-of-the-art algorithms. 
For example, in the Movie Lens dataset ($n$=4,000, $d$=6,000), \algname is 20$\times$ faster than the next best algorithm while returning the same answer. 
% We empirically verify the advantages of \algname on several high-dimensional real-world datasets. \algname runs significantly faster than existing approaches and returns the same solution.
\algname requires no preprocessing of the data and includes a hyperparameter that practitioners may use to trade off accuracy and runtime.
We also propose a variant of our algorithm, named \algnamenospace-$\alpha$, which achieves further speedups by employing non-uniform sampling across coordinates. 
Finally, we demonstrate how known preprocessing techniques can be used to further accelerate \algnamenospace, and discuss applications to Matching Pursuit and Fourier analysis. 
\end{abstract}

\input{2-introduction}

\input{3-related_work}

\input{4-preliminaries}

\input{5-algorithm}

\input{6-theory}

\input{7-experiments}

\input{8-conclusions_and_limitations}

\bibliographystyle{plain}
\bibliography{biblio}

\input{9.1-appendix_additional_related_work}
\input{9.2-appendix_proofs}

\input{9.3-appendix_datasets}
\input{9.4-appendix_additional_experiments}
\input{9.5-appendix_preprocessing}
\input{9.6-appendix_high_dim}

\input{9.7-appendix_symmetric}

\end{document}

%% file: _1-defs.tex
\newtheorem{theorem}{Theorem}

\newcounter{numcount}
\newif\iflong
\longfalse

\newif\ifdraft
\drafttrue

%% file: 2-introduction.tex
% !TEX root = 0-main.tex

\section{Introduction}
\label{sec:intro}

\let\thefootnote\relax\footnotetext{$*$ and $\#$ denote equal contribution.}
\let\thefootnote\relax\footnotetext{Correspondence should be addressed to M.T. (\texttt{motiwari@stanford.edu})}
\let\thefootnote\relax\footnotetext{1: Department of Computer Science, Stanford University}
\let\thefootnote\relax\footnotetext{2: Mathematical Institute, Oxford University}
\let\thefootnote\relax\footnotetext{3: Department of Computer Science, University College London}
\let\thefootnote\relax\footnotetext{4: Electrical and Computer Engineering, University of Illinois at Urbana-Champaign}
\let\thefootnote\relax\footnotetext{5: Department of Epidemiology, Harvard T.H. Chan School of Public Health}

% \textcolor{red}{[Martin: I suggest the following organization for this section: 
% \begin{enumerate}
%     \item Paragraph 1 describes MIPS and its wide use. Mention the name (but without details) of related applications embedding-based or matrix factorization-based recommender systems, Matching Pursuit, or arises in the inference stages of many other applications. 
%     \item Paragraph 2 describes the issue (computational cost) and the state-of-art efforts to address this issue. Postpone the details to the Related work subsection.
%     \item Paragraph 3 describes our algorithm and summarizes our contribution, including algorithm, theoretical results, and expirical results. 
% \end{enumerate}
% ]}
%\martin{[Paper org]  para1 of introduction describes the problem formulation at a high-level. s1: what is MIPS and where can it be used (with references). s2: why it is desirable to accelerate MIPS (with reference),  s3: more details about parameters $n$ and $d$, including math formulation, usual values for these two parameters, and which one do we care more about) }

The Maximum Inner Product Search problem (MIPS) \cite{shrivastavaAsymmetricLSHALSH2014,neyshaburSymmetricAsymmetricLSHs2015,yuGreedyApproachBudgeted2017} is a ubiquitous task that arises in many machine learning applications, such as matrix-factorization-based recommendation systems \cite{korenMatrixFactorizationTechniques2009,cremonesi2010performance}, multi-class prediction \cite{deanFastAccurateDetection2013,jainActiveLearningLarge2009}, structural SVM \cite{joachims2006training,joachims2009cutting}, and computer vision \cite{deanFastAccurateDetection2013}. 
Given a $\textit{query}$ vector $\mathbf{q} \in \mathbb{R}^d$ and $n$ $\textit{atom}$ vectors $\mathbf{v}_1, \ldots, \mathbf{v}_n \in \mathbb{R}^d$, MIPS aims to find the atom most similar to the query:
\begin{equation}
\label{eqn:mips}
    i^* = \argmax_{i \in \{1,\cdots,n\}} \mathbf{v}_i^T \mathbf{q}
\end{equation}
For example, in recommendation systems, the query $\mathbf{q}$ may represent a user and the atoms $\mathbf{v}_i$'s represent items with which the user can interact; MIPS finds the best item for the user, as modeled by their concordance $\mathbf{v}_i^T \mathbf{q}$ \cite{amagata_reverse_2021, aouali_reward_2022}.
In many applications, the number of atoms $n$ and the feature dimension $d$ can easily be in the millions, so it is critical to solve MIPS accurately and efficiently \cite{hirata_solving_2022}.

The na\"ive approach evaluates all $nd$ elements and scales as $O(nd)$.
Most recent works focus on reducing the scaling with $n$ and scale at least linearly in $d$ \cite{lorenzen_revisiting_2021-1}, which may be prohibitively slow in high-dimensional settings.
\cite{liuBanditApproachMaximum2019} proposed a sampling-based approach that improved the complexity to $O(n\sqrt{d})$.
In this work, we focus on further improving the complexity with respect to $d$ and providing a tunable hyperparameter that governs the tradeoff between accuracy and speed, a need identified by previous works \cite{yuGreedyApproachBudgeted2017}.

We propose \algnamenospace, a new randomized algorithm for MIPS whose complexity is independent of $d$.
We provide theoretical guarantees that \algname recovers the exact solution to Equation \eqref{eqn:mips} with high probability in $\tilde{O}(\frac{n}{\Delta^2})$\footnote{The $\tilde{O}$ notation hides logarithmic factors.} time, where $\Delta$ is an instance-specific factor that does not depend on $d$.
We have also performed comprehensive experiments to evaluate our algorithm's performance in two synthetic and two real-world datasets. 
For example, in the Movie Lens dataset ($n = 4,000$, $d=6,000$) \cite{movie2015}, \algname is 20$\times$ faster than prior state-of-the-art while returning the same answer.

At a high-level, instead of computing the inner product $\mathbf{v}_i^T \mathbf{q}$ for each atom $i$ using all $d$ coordinates, \algname estimates them by subsampling a subset of coordinates. Since more samples give higher estimation accuracy, \algname adaptively samples more coordinates for top atoms to discern the best atom. The specific adaptive sampling procedure is motivated by multi-armed bandits (MAB) \cite{even-darActionEliminationStopping2006}.

% \textbf{Contributions:} We propose a new algorithm for MIPS called \algnamenospace. \algname is a randomized algorithm that recovers the exact solution to Equation \eqref{eqn:mips} with high probability in $\tilde{O}(\frac{n}{\Delta^2})$\footnote{The $\tilde{O}$ notation hides logarithmic factors.} time, where $\Delta$ is the average coordinate-wise gap between the best and second best inner product.
% Under general assumptions about the data generating distribution, its runtime not explicitly scale with $d$. 
% Furthermore, \algname provides a tunable hyperparameter to trade off accuracy and runtime, enables users to provide a computational budget to control the runtime and allows for finding $\epsilon$-suboptimal solutions.
% Unlike prior work, \algname requires no preprocessing or normalization of the data and is easily parallelizable. 

% TODO -- Sample without replacement
% TODO: -- We have lower complexity
\algname is easily parallelizable and can be used with other optimization objectives that decompose coordinate-wise.
Unlike previous works, it does not require preprocessing or normalization of the data, nor does it require the query or atoms to be nonnegative \cite{yuGreedyApproachBudgeted2017}. 
\algname also has a tunable hyperparameter to trade off accuracy and speed.
We also developed several extensions of \algnamenospace.
First, we propose \algnamenospace-$\alpha$, which provides additional runtime speedups by sampling coordinates intelligently in Section \ref{subsec:additional_speedup_techniques}.
Second, we extend \algname to find the $k$ atoms with the highest inner products with the query ($k$-MIPS) in our experiments in Section \ref{sec:experiments} and Appendix \ref{app:additional_experiments}.
Third, we discuss how \algname can be used in conjunction with preprocessing techniques in Appendix \ref{app:preprocessing} and examples of downstream applications in Appendix \ref{app:high_dim}.

% \algname generalizes to finding the $k$ atoms with the highest inner products with the query (k-MIPS), can be used with other optimization objectives that decompose coordinate-wise, and does \textit{not} require the query or atoms to be nonnegative as in prior work such as \citet{yuGreedyApproachBudgeted2017}.
% We provide theoretical guarantees about the performance of \algname in Section \ref{sec:theory} and validate our claims experimentally in Section \ref{sec:experiments}.
% Additionally, we propose \algnamenospace-$\alpha$, which provides additional runtime speedups by sampling coordinates intelligently.
% Our experiments also demonstrate that our algorithms' scaling with $d$ is superior to existing state-of-the-art approaches.%, which justifies our assumption that $\Delta$ is often independent of $d$. 
% We also discuss how \algname can be used in conjunction with preprocessing techniques in Appendix \ref{app:preprocessing} and for example downstream applications in Appendix \ref{app:high_dim}.

%The requirement for the ability to provide a fixed computational budget or a hyperparameter that governs the tradeoff between accuracy and runtime has been identified by prior work as important in applications . 

%We also provide an open-source implementation of \algnamenospace, \algnamenospace-$\alpha$,  and other baseline MIPS algorithms that may be of independent interest.

%% file: 3-related_work.tex
% \section{Related Work}
\subsection{Related work}
\label{sec:relatedwork}

% \martin{[Paper org] 1. Move this section as subsection 1.1. Keep this part concise, maybe 1-1.5 pages. Consider the following structure: 
% \begin{enumerate}
%     \item Para1 talks about applications of MIPS and why it's desirable to accelerate it, to accelerate it in wrt $d$ but not $n$, and to have a tunable parameter for speed and accuracy. This mostly contains the current 2.1 and 2.2.
%     \item Para2 talk about existing popular methods to accelerate MIPS. Each part should consist 1-3 sentences and briefly describe the method (with reference) and the limitations. 1. product quantization, 2. NN-based methods, 3. LSH-based methods. 4. promixity graphs-based approaches. This mostly contains the first few paragraphs of the current 2.3. 
%     \item Para3 talks about methods more related to our work, such as BoundedME. \item Para4 talks about best-arm identification. 
% \end{enumerate}
% }

% Uses of MIPS

% \subsection{Applications of MIPS}
%\martin{This is good stuff. Consider move some of these contents to Intro para1 to communicate that 1. MIPS is widely-used, 2. scaling up MIPS are important. It is desirable to cite many papers in Intro para1 to make the message stronger.}

\textbf{MIPS applications:} MIPS arises naturally in many information retrieval contexts \cite{sivic_video_2003, dong2012duplicate, BoytsovNMN16} and for augmenting large, auto-regressive language models \cite{borgeaudImprovingLanguageModels2022a}. 
%For example, MIPS has recently been used to improve large, auto-regressive language models by conditioning on document chunks retrieved from a corpus of trillions of tokens, effectively augmenting models with massive-scale memory \cite{borgeaudImprovingLanguageModels2022a}.
MIPS is also a subroutine in the Matching Pursuit problem (MP) and its variants, such as Orthogonal Matching Pursuit (OMP) \cite{locatelloUnifiedOptimizationView2017}.
%In these problems, the query vector is approximated via a sparse decomposition into the dictionary atoms; the canonical formulations of MP and OMP can be viewed as iterated versions of MIPS.
MP and other iterative MIPS algorithms have found a many applications, e.g., to find a sparse solution of underdetermined systems of equations \cite{donohoSparseSolutionUnderdetermined2012} and accelerate conditional gradient methods \cite{song_accelerating_2022, xu_breaking_2021}.
MIPS also arises in the inference stages of many other applications, such as for deep-learning based multi-class or multi-label classifiers \cite{deanFastAccurateDetection2013, jainActiveLearningLarge2009} and has been used as a black-box subroutine to improve the learning and inference in unnormalized log-linear models when computing the partition function is intractable \cite{mussmannLearningInferenceMaximum2016}.%, e.g., in Markov Random Fields or multinomial logistic regression 
% \subsection{Large datasets and tunability}
% \martin{I find this aspect less impressive. If you want to include it, it is important to clearly explain its importance in Intro para1 or para2}
% Recent advances in computational power and digital data collection have created a need for efficient MIPS algorithms on huge datasets.
% In real-world datasets, $n$ and $d$ are often in the millions \cite{simhadri_results_2022}.
% The Netflix Prize dataset, for example, contains over 17,000 movies and 480,000 users ratings.
% \subsection{Existing Approaches}
% APPROX
%Much existing work focuses on an approximate solution to the MIPS problem when it may computationally intractable to solve exactly.
%Most approaches focus improving MIPS algorithms' scaling with $n$, the number of available atoms. 
%Instead, we focus on the improving the scaling with $d$, the dimensionality of the vectors, to enable MIPS on very high dimensional datasets, such as time series.
%While a naive solution to Equation \eqref{eqn:mips} would be to compute all dot products explicitly in $O(nd)$ time, it is sufficient to determine a \textit{ranking} of the inner products without necessarily computing them explicitly.

\textbf{MIPS algorithms:} Many approaches focus on solving approximate versions of MIPS. 
Such work often assumes that the vector entries are nonnegative, performs non-adaptive sampling \cite{luSamplingApproximateMaximum2017,ballardDiamondSamplingApproximate2015, lorenzen_revisiting_2021-1, ding_fast_2019, yuGreedyApproachBudgeted2017}, or rely on product quantization \cite{dai_norm-explicit_2020, wu_local_2019, guo_accelerating_2020, guo_new_2019, matsui_invited_2018, douze_polysemous_2016, ge_optimized_2013, babenko_inverted_2012, jegou_searching_2011, jegou_product_2011}.
Many of these algorithms require significant preprocessing, are limited in their adaptivity to the underlying data distribution, provide no theoretical guarantees, or scale linearly in $d$---all drawbacks that have been identified as bottlenecks for MIPS in high dimensions \cite{ponomarenko_comparative_2014}.

A large family of MIPS algorithms are based on locality-sensitive hashing (LSH) \cite{indykApproximateNearestNeighbors1998,shrivastavaAsymmetricLSHALSH2014,shrivastavaImprovedAsymmetricLocality2015,neyshaburSymmetricAsymmetricLSHs2015,huangQueryawareLocalitysensitiveHashing2015,song_promips_2021, lu_adalsh_2021, shrivastava_improved_2014, wu_h2sa-alsh_2022, huang_accurate_2018, ma_learning_2021, NIPS2015_2823f479, yan_norm-ranging_2018}.
A shortcoming of these LSH-based approaches is that, in high dimensions, the maximum dot product is often small compared to the vector norms, which necessitates many hashes and significant storage space (often orders of magnitude more than the data itself). 
%Our experiments in Section \ref{sec:experiments}, where we compare \algname directly to several representative LSH-based approaches, are consistent with this observation.
%w\martin{Is this right that LSH can be viewed as a non-adaptive sampling version of our algorithm?}
Many other MIPS approaches are based on proximity graphs, such as ip-NSW \cite{morozov_non-metric_2018} and related work \cite{liu_understanding_2020, feng_reinforcement_2023, tan_efficient_2019, tan_norm_2021, zhou_mobius_2019, chen_finger_2022, zhang_grasp_2022, alexander_approximate_2011, Malkov2016a, MalkovPLK14}. 
These approaches use preprocessing to build an index data structure that allows for more efficient MIPS solutions at query time.
However, these approaches also do not scale well to high dimensions as the index structure (an approximation to the true proximity graph) breaks down due to the curse of dimensionality \cite{liu_understanding_2020}.

Perhaps most similar to our work is BoundedME \cite{liuBanditApproachMaximum2019}.
Similar to our method, their approach presents a solution to MIPS based on adaptive sampling but scales as $O(n\sqrt{d})$, worse than our algorithm that does not scale with $d$.
%\martin{include both $n$ and $d$ when talking about scaling to avoid confusion}
This is because in BoundedME, the number of times each atom is sampled is predetermined by $d$ and not adaptive to the actual \textit{values} of the sampled inner products; rather, is only adaptive to their relative \textit{ranking}. 
Intuitively, this approach is wasteful because information contained in the sampled inner product's values is discarded. 
%Instead, \algname is more adaptive to the underlying data distribution and uses information about the estimated inner products to determine what atoms to sample next.
%The runtime of \algname is determined by the \textit{hardness} of the problem, i.e., the gap between the highest inner product and second-highest inner product.
%When this gap does not depend on $d$, as is the case under general assumptions, \algnamenospace's complexity scales as $O(1)$ with respect to $d$ and is independent of the data dimension.
%We find that \algname is $12\times$ faster than theirs even on modest datasets such as the Netflix Prize dataset. Furthermore, we provide a modification to \algnamenospace, dubbed \algnamenospace-$\alpha$, that provides a further $2\times$ speedup over \algname (a $24\times$ over \texttt{BoundedME}) by sampling coordinates intelligently

Additional related work is discussed in Appendix \ref{app:add_related_work}.
% \martin{This is an interesting explanation. I am not sure if this is the correct intuition. please double check.}
% \Jey{The original Median Elimination algorithm also uses ranking of the atoms, but its sample complexity is $\frac{2n}{{\epsilon^2}} \log {\frac{1}{\delta}}$.}
%\mo{Downplay connection to MABs}

\textbf{Multi-armed bandits:} 
% We also briefly recapitulate the problem of best arm identification from multi-armed bandits \cite{even-darActionEliminationStopping2006, jamiesonBestarmIdentificationAlgorithms2014}.
\algname is motivated by the best-arm identification problem in multi-armed bandits \cite{even-darActionEliminationStopping2006, karninAlmostOptimalExploration2013, audibertBestArmIdentification2010, jamiesonBestarmIdentificationAlgorithms2014, jamiesonLilUCBOptimal2014, jamiesonNonstochasticBestArm2016, bubeckPureExplorationFinitelyarmed2011, bardenetConcentrationInequalitiesSampling2015, boucheronConcentrationInequalitiesNonasymptotic2013, even-darPACBoundsMultiarmed2002,kalyanakrishnanPACSubsetSelection2012}. In a typical setting, we have $n$ arms each associated with an expected reward $\mu_i$. At each time step $t = 0,1,\cdots,$ we decide to pull an arm $A_t\in \{1,\cdots,n\}$, and receive a reward $X_t$ with $E[X_t] = \mu_{A_t}$. The goal is to identify the arm  with the largest reward with high probability while using the fewest number of arm pulls.
% Many algorithms have been designed for the best-arm identification problem \cite{even-darActionEliminationStopping2006, karninAlmostOptimalExploration2013, audibertBestArmIdentification2010, jamiesonBestarmIdentificationAlgorithms2014, jamiesonLilUCBOptimal2014, jamiesonNonstochasticBestArm2016, bubeckPureExplorationFinitelyarmed2011, bardenetConcentrationInequalitiesSampling2015, boucheronConcentrationInequalitiesNonasymptotic2013, even-darPACBoundsMultiarmed2002,kalyanakrishnanPACSubsetSelection2012}. 
% In its reduction to a best-arm identification problem, \algname reframes each atom as an arm; pulling an arm then corresponds to evaluating a single, \textit{coordinate-wise} product; see Table \ref{table:reduction}. We emphasize that our work does not propose a novel multi-armed bandit algorithm; rather, we apply adaptive sampling techniques to MIPS. This reduction of computationally intensive machine learning problems to statistical problems based on adaptive sampling has demonstrated success in other areas \cite{tiwari2020banditpam, bagaria2018medoids, bagaria2018adaptive, zhangAdaptiveMonteCarlo2019, bagaria2021bandit}.
The use of MAB-based adaptive sampling to develop computationally efficient algorithms has seen many applications, such as random forests and $k$-medoid clustering \cite{tiwari2020banditpam, bagaria2018medoids, bagaria2018adaptive, zhangAdaptiveMonteCarlo2019, bagaria2021bandit}.

%% file: 4-preliminaries.tex
% !TEX root = 0-main.tex

\section{Preliminaries and Notation}
\label{sec:preliminaries}

We consider a query $\mathbf{q} \in \mathbb{R}^d$ and $n$ atoms $\mathbf{v}_1, \ldots, \mathbf{v}_n \in \mathbb{R}^d$. 
Let $[n]$ denote $\{1,\ldots,n\}$, $q_j$ the $j$th element of $\mathbf{q}$, and $v_{ij}$ the $j$th element of $\mathbf{v}_i$. 
% \iscomment{$j$-th element of $\mathbf{v}_i$?}
For a given query $\mathbf{q} \in \mathbb{R}^d$, the MIPS problem is to find the solution to Equation \eqref{eqn:mips}: $i^* = \argmax_{i \in [n]}  \mathbf{v}_i^T \mathbf{q}$.

We let $\mu_i \coloneqq \frac{\mathbf{v_i}^T \mathbf{q}}{d}$ denote the \textit{normalized inner product} for atom $\mathbf{v}_i$. Since the inner products $\mathbf{v_i}^T \mathbf{q}$ tend to scale linearly with $d$ (e.g., if each coordinate of the atoms and query are drawn i.i.d.), each $\mu_i$ should not scale with $d$. 
%\iscomment{not clear what ``does not scale with $d$'' means exactly. Maybe ``Since inner products $\mathbf{v_i}^T \mathbf{q}$ tend to scale linearly with $d$, $\mu_i$'s do not scale with $d$''}
Note that $\argmax_{i \in [n]} \mathbf{v}_i^T \mathbf{q} = \argmax_{i \in [n]} \mu_i$ so it is sufficient to find the atom with the highest $\mu_i$.
Furthermore, for $i \neq i^*$ we define the gap of atom $i$ as $\Delta_i \coloneqq \mu_{i^*} - \mu_i \geq 0$ and the minimum gap as $\Delta \coloneqq \min_{i \neq i^*}{\Delta_i}$. 
We primarily focus on the computational complexity of MIPS with respect to $d$.
%As described in Section \ref{sec:intro}, a na\"ive exhaustive approach that computes the inner product of each $\mathbf{v}$ with the query $\mathbf{q}$ incurs computational cost $O(nd)$.
% Other recent approaches improve the scaling with $d$ to $O(n\sqrt{d})$ \cite{liuBanditApproachMaximum2019}.
% In the rest of the paper, we describe \algnamenospace, which scales as $\tilde{O}(\frac{n}{\Delta^2})$, i.e., $O(1)$ with respect to $d$, under reasonable assumptions on $\Delta$.

%% file: 5-algorithm.tex
% !TEX root = 0-main.tex

\section{Algorithm}
\label{sec:algo}

\begin{table}
\caption{MIPS as a best-arm identification problem.}
\label{alg:formulation}
\centering
\vspace{1em}
\begin{tabular}{p{2.3cm}p{5cm}p{5cm}}  % width
\toprule
Terminology & Best-arm identification & MIPS \\
\midrule
Arms & $i=1,\ldots,n$ & Atoms $\mathbf{v}_1,\ldots,\mathbf{v}_n$ \\
Arm parameter $\mu_i$ & Expected reward $\mathbb{E}[X_i]$ & Average coordinate-wise product $\frac{\mathbf{v}i^T \mathbf{q}}{d}$ \\
Pulling arm $i$ & Sample a reward $X_i$ & Sample a coordinate $J$ with reward $q_J v_{iJ}$ \\
Goal & Identify best arm with probability at least $1-\delta$ & Identify best atom with probability at least $1-\delta$ \\
\bottomrule
\end{tabular}
\end{table}

% \begin{table*}[t]
% \label{table:reduction}

% \centering
% \begin{tabular}{|c|c|c|}
% \hline
% \textbf{Terminology} & \textbf{Best-arm identification} & \textbf{MIPS} \\ \hline
% Arms & $i=1,\ldots,n$ & Atoms $\mathbf{v}_1,\ldots,\mathbf{v}_n$  \\ 
% Arm parameter $\mu_i$ & Expected reward $\mathbb{E}[X_i]$ & Average coordinate-wise product $\frac{\mathbf{v}_i^T \mathbf{q}}{d}$ \\ 
% Pulling arm $i$ & Sample a reward $X_i$ & Sample a coordinate $J$ with reward $q_J v_{iJ}$ \\ 
% Goal & Identify best arm with probability at least $1-\delta$ & Identify best atom with probability at least $1-\delta$\\ 
% \hline
% \end{tabular}
% \caption{MIPS as a best-arm identification problem.}
% \label{alg:formulation}
% \end{table*}

\begin{algorithm}[h]
\caption{\algnamenospace}
\label{alg:bandit_based_search}
\textbf{Input}: Atoms $\mathbf{v}_1, \dots, \mathbf{v}_n \in \mathbb{R}^d$, query $\mathbf{q} \in \mathbb{R}^d$, error probability $\delta$, sub-Gaussian parameter $\sigma$
\textbf{Output}: $i^* = \argmax_{i \in [n]} \mathbf{q}_i^T \mathbf{v}$
\begin{algorithmic}[1]
\State $\mathcal{S}_{\text{solution}} \leftarrow [n]$
\State $d_{\text{used}} \leftarrow 0$
\State For all $i \in \mathcal{S}_{\text{solution}}$, initialize $\hat{\mu}_i \leftarrow 0$, $C_{d_\text{used}} \leftarrow \infty$
\While{$d_{\text{used}} < d$ and $|\mathcal{S}_{\text{solution}}| > 1$}
\State Sample a new coordinate $J \sim \text{Unif}[d]$
    \ForAll{$i \in \mathcal{S}_{\text{solution}}$}
        \State $\hat{\mu}_i \leftarrow \frac{ d_{\text{used}} \hat{\mu}_i + v_{iJ} q_J}{ d_{\text{used}} + 1 }$
        \State $\left(1-\frac{\delta}{2 n d_\text{used}^2}\right)$-CI: $C_{d_\text{used}} \gets \sigma \sqrt{  \frac{ 2 \log \left(4nd^2_\text{used} / \delta \right) }{ d_{\text{used}} + 1} }$ 
    \EndFor
\State $\mathcal{S}_{\text{solution}} \leftarrow \{i : \hat{\mu}_i + C_{d_\text{used}} \geq \max_{i'} \hat{\mu}_{i'} - C_{d_\text{used}}\}$ 
\State $d_{\text{used}} \leftarrow d_{\text{used}} + 1$
\EndWhile
\State If $\vert \mathcal{S}_{\text{solution}} \vert > 1$, update $\hat{\mu}_i$ to be the exact value $\mu_i = \mathbf{v}_i^T q$ for each atom in $\mathcal{S}_{\text{solution}}$ using all $d$ coordinates
\State \textbf{return} $i^* = \argmax_{i \in \mathcal{S}_{\text{solution}}} \hat{\mu}_i$
\end{algorithmic}
\end{algorithm}

The \algname algorithm is described in Algorithm \ref{alg:bandit_based_search} and is motivated by best-arm identification algorithms. As summarized in Table \ref{alg:formulation}, we can view each atom $\mathbf{v}_i$ as an arm with the arm parameter $\mu_i \coloneqq \frac{\mathbf{v}_i^T\mathbf{q}}{d} $. When pulling an arm $i$, we randomly sample a coordinate $J \sim \text{Unif}[d]$ and evaluate the inner product at the coordinate as $X_i = q_J v_{iJ} $.
%The $d$ term ensures that $X$ is an unbiased estimator of the arm parameter $\mathbb{E}[X] = \mu_i$. 
% This reformulation of MIPS is summarized in Table \ref{alg:formulation}.
% Solvers
Using this reformulation, the best atom can be estimated using techniques from best-arm algorithms. 
% like the Upper Confidence Bound (UCB) algorithm \cite{lai1985asymptotically} and successive elimination \cite{successiveelimination}.

\algname can be viewed as a combination of UCB and successive elimination \cite{lai1985asymptotically,successiveelimination,zhang2019adaptive}.
Algorithm \ref{alg:bandit_based_search} uses the set $\mathcal{S}_{\text{solution}}$ to track all potential solutions to Equation \eqref{eqn:mips}; $\mathcal{S}_{\text{solution}}$ is initialized as the set of all atoms $[n]$. We will assume that, for a fixed atom $i$ and a randomly sampled coordinate, the random variable $X_i = q_J v_{iJ}$ is $\sigma$-sub-Gaussian for some known parameter $\sigma$. 
% \iscomment{$\sigma_i$ doesn't appear in algorithm. Is it $c_0$?}. 
With this assumption,  Algorithm \ref{alg:bandit_based_search} maintains a mean objective estimate $\hat{\mu}_i$ and confidence interval (CI) for each potential solution $i\in \mathcal{S}_{\text{solution}}$, where the CI depends on the error probability $\delta$ as well as the sub-Gaussian parameter $\sigma$. We discuss the sub-Gaussian parameter and possible relaxations of this assumption in Subsections \ref{subsec:subgaussianity} and \ref{subsec:littledelta}.

%- Assume loss function decomposes over coordinates, as is the case for $L_1$ or $L_2$ loss

\subsection{Additional speedup techniques}
\label{subsec:additional_speedup_techniques}

% We present several techniques to further improve the computational efficiency of BanditMIPS. 

\textbf{Non-uniform sampling reduces variance:} In the original version of \algnamenospace, we sample a coordinate $J$ for all atoms in $\mathcal{S}_{\text{solution}}$ uniformly from the set of all coordinates $[d]$.
However, some coordinates may be more informative of the inner product than others.
For example, larger entries of $\mathbf{v}_i$ may contribute more to the inner product with $\mathbf{q}$.
% Procedure
As such, we sample each coordinate $j \in [d]$ with probability $w_j \propto q_j ^ {2\beta}$
% \iscomment{shouldn't we include the $q_j^2$ in here too?}
and $\sum_j w_j=1$, and estimate the arm parameter $\mu_i$ of atom $i$ as $X = \frac{1}{w_J} q_J v_{iJ}$. 
% Property
$X$ is an unbiased estimator of $\mu_i$ and the specific choice of coordinate sampling weights minimizes the combined variance of $X$ across all atoms; different values of $\beta$ corresponds to the minimizer under different assumptions.
We provide theoretical justification of this weighting scheme in Section \ref{sec:theory}.
We note that the effect of this non-uniform sampling will only accelerate the algorithm.
%We leave an analysis of its exact affect on the big-$O$ complexity of the algorithm to future work. \martin{Was this sentence required by the reviewers? How about "We empirically verified that this approach substantially improved speed and will leave the theoretical analysis to future work"}

\textbf{Warm start increases speed:} One may wish to perform MIPS for a batch of $m$ queries instead of just a single query, solving $m$ separate MIPS problems. In this case, we can cache the atom values for all atoms across a random subset of coordinates, and provide a warm start to BanditMIPS by using these cached values to update arm parameter estimates $\hat{\mu}_i$, $C_i$, and $\mathcal{S}_{\text{solution}}$ for all $m$ MIPS problems. Such a procedure will eliminate the obviously less promising atoms and avoid repeated sampling for each of the $m$ MIPS problems and increases computational efficiency. We note that, since the $m$ MIPS problems are independent, the theoretical guarantees described in Section \ref{sec:theory} still hold across all $m$ MIPS problems simultaneously. 
%the dependency of this first set of arm pulls across the $m$ MIPS problems does not invalidate the theoretical guarantees.

\subsection{Sub-Gaussian assumption and construction of confidence intervals}
\label{subsec:subgaussianity}

%\subsubsection{Sampling:}
%In our implementation, we sample coordinates in batches instead of one-by-one.
% For the purposes of the theoretical analysis of Algorithm \ref{alg:bandit_based_search}, we assumed that we sample with replacement and, if more than $d$ coordinates are sampled for a given atom, then we perform an exact computation of that atom's inner product with the query vector (incurring another $d$ computations for a maximum possible $2d$ sampled coordinates).
% In our implementation, however, we sample without replacement, in line with \citet{liuBanditApproachMaximum2019}. We find that this does not materially affect our results.

%\subsubsection{Construction of CIs:}

Crucial to the accuracy of Algorithm \ref{alg:bandit_based_search} is the construction of of the $(1 - \delta)$-CI based on the $\sigma$-sub-Gaussianity of each $X_i = q_J v_{iJ}$.
We note that the requirement for $\sigma$-sub-Gaussianity is rather general. 
In particular, when the coordinate-wise products between the atoms and query are bounded in $[a, b]$, then each $X_i$ is $\frac{b^2 - a^2}{4}$-sub-Gaussian.
This is commonly the case, e.g., in recommendation systems where user ratings (each element of the query and atoms) are integers between 0 and 5, and we use this implied value of $\sigma$ in our experiments in Section \ref{sec:experiments}.

%Important to the \algname algorithm is the estimation of the sub-Gaussian parameters $\sigma$.
%When the data is bounded in the range $[a, b]$, as is the case in all of our datasets, each arm is $(b^2 - a^2) / 4$-sub-Gaussian. 
The $\frac{b^2 - a^2}{4}$-sub-Gaussianity assumption allows us to compute $1 - \delta$ CIs via Hoeffding's inequality, which states that for any random variable $S_n = Y_1 + Y_2 + \ldots Y_n$ where each $Y_i \in [a, b]$
\begin{equation*}
P(|S_n - \mathbb{E}[S_n]| > \epsilon) \leq \text{exp}\left(\frac{-2\epsilon^2}{n(b - a)^2}\right).
\end{equation*}

Setting $\delta$ equal to the right hand side and solving for $\epsilon$ gives the width of the confidence interval.
$\sigma = \frac{b^2 - a^2}{4}$ acts as a variance proxy used in the creation of the confidence intervals %\iscomment{not sure what ``variance proxy for confidence intervals'' means} 
by \algnamenospace; smaller variance proxies should result in tighter confidence intervals and lower sample complexities and runtimes. 

In other settings where the sub-Gaussianity parameter may not be known \textit{a priori}, it can be estimated from the data or the CIs can be constructed using the empirical Bernstein inequality instead \cite{maurerEmpiricalBernsteinBounds2009}.

%% file: 6-theory.tex
\section{Theoretical Analysis}
\label{sec:theory}

% In this section, we present theoretical results on the correctness and computational complexity of BanditMIPS.
% We also, discuss the effect of the hyperparamater $\delta$ and the necessary assumptions about the arm gaps (the $\Delta_i$'s).
% Finally, we derive the optimal weights for non-uniformly sampling the coordinates. 

\label{sec:analysis}
\textbf{Analysis of the Algorithm:}
%For a given query $\mathbf{q}$, let $i^* = \argmax_{i \in [n]} \mu_{i}$ be the best atom and $\Delta_{i} \coloneqq \mu_{i^*} - \mu_{i}$ be the gap between other atoms and the best atom, as in Section \ref{sec:preliminaries}. \martin{No need to define the same thing again.}
For Theorem \ref{thm:specific}, we assume that, for a fixed atom $\mathbf{v}_i$ and $d_\text{used}$ randomly sampled coordinates, the $(1-\delta')$ confidence interval scales as $C_{d_\text{used}}(\delta') = O\left(\sqrt{\frac{\log 1 / \delta'}{d_\text{used}}}\right)$ (note that we use $d_\text{used}$ and $\delta'$ here because we have already used $d$ and $\delta$).
We note that the sub-Gaussian CIs satisfy this property, as described in Section \ref{subsec:subgaussianity}.
%Please see more details in below.

% In this section, we prove that \algname returns the optimal atom for a node split with high probability. Furthermore, we provide bounds on computational complexity of \algnamenospace, leading to the logarithmic dependency on the dimensionality of the problem $d$.

% We provide theoretical results Consider the MIPS problem, Equation \ref{eqn:mips}, with query vector $\mathbf{q}$ and $n$ atoms $\mathbf{v}_1, \ldots,\mathbf{v}_n$.
% Suppose $i^* = \argmin_{i \in [n]} \mu_{i}$ is the index of the atom with the highest-inner product with $\mathbf{q}$.
% For any other atom $\mathbf{v}_i$, define $\Delta_{i} \coloneqq \mu_{i^*} - \mu_{i}$.
% To state the following results, we will assume that, for a fixed atom $\mathbf{v}_i$ and $d'$ randomly sampled coordinates, the $(1-\delta)$ confidence interval scales as $C_{i}(d', \delta) = O(\sqrt{\frac{\log 1/\delta}{d'}})$. 
% We discuss how to construct such confidence intervals with this property \todo{Appendix link} \cite{van2000asymptotic}.
% With this assumption, we state the following theorem:

\begin{theorem}
\label{thm:specific}
Assume $\exists~c_0 > 0$ s.t. $\forall~\delta'>0$, $d_\text{used}>0$, $C_{d_\text{used}}(\delta') < c_0\sqrt{\frac{\log 1 / \delta'}{d_\text{used}}}$.
% \iscomment{update?} 
With probability at least $1-\delta$, \algname returns the correct solution to Equation \eqref{eqn:mips} and uses a total of $M$ computations, where
\begin{align} 
\label{eqn:instance_bd}
M \leq \sum_{i \in [n]}  \min \left[ \frac{16c_0^2}{\Delta_{i}^2} \log \left( \frac{n}{\delta \Delta_i} \right) + 1, 2d \right].
\end{align}
%\martin{The standard argument is an upper bound on $E[M]$, because the number of pulls may be very large when the CIs do not always hold.}
\end{theorem}

Theorem \ref{thm:specific} is proven in the appendices.
We note that $c_0$ is the sub-Gaussianity parameter described in Section \ref{subsec:subgaussianity} and is a constant.
Intuitively, Theorem \ref{thm:specific} states that with high probability, \algname returns the atom with the highest inner product with $\mathbf{q}$.
The instance-wise bound Equation \eqref{eqn:instance_bd} suggests the computational cost of a given atom $\mathbf{v}_i$, i.e., $\min \left[ \frac{16c_0^2}{\Delta_{i}^2} \log\left(\frac{n }{\delta \Delta_i}\right) + 1, 2d \right]$, depends on $\Delta_{i}$, which measures how close its optimization parameter $\mu_{i}$ is to $\mu_{i^*}$. 
Most reasonably different atoms $i\neq i^*$ will have a large $\Delta_{i}$ and incur an $O\left(\frac{1}{\Delta^2}\log \frac{n}{\delta \Delta_i}\right)$ computation that is independent of $d$ when $d$ is sufficiently large.

Important to Theorem \ref{thm:specific} is the assumption that we can construct $(1-\delta')$ CIs $C_{i}(d_\text{used}, \delta')$ that scale as $O(\sqrt{\frac{\log 1/\delta'}{d_\text{used}}})$.
As discussed in Section \ref{subsec:subgaussianity}, this is under general assumptions, for example when the estimator $X_i = q_J v_{iJ}$ for each arm parameter $\mu_i$ has finite first and second moments \cite{catoni2012challenging} or is bounded.

%\todo{Add translation to wall-clock time}
Since each coordinate-wise multiplication only incurs $O(1)$ computational overhead to update running means and confidence intervals, sample complexity bounds translate directly to wall-clock times bounds up to constant factors. For this reason, our approach of focuses on sample complexity bounds, in line with prior work \cite{tiwari2020banditpam, bagaria2018adaptive}.

\label{subsec:littledelta}
\textbf{Discussion of the hyperparameter $\delta$:}
The hyperparameter $\delta$ allows users to trade off accuracy and runtime when calling Algorithm \ref{alg:bandit_based_search}.
A larger value of $\delta$ corresponds to a lower error probability, but will lead to longer runtimes because the confidence intervals constructed by Algorithm \ref{alg:bandit_based_search} will be wider and atoms will be filtered more slowly. 
Theorem \ref{thm:specific} provides an analysis of the effect of $\delta$ and in Section \ref{sec:experiments}, we discuss appropriate ways to tune it. 
We note that setting $\delta = 0$ reduces Algorithm \ref{alg:bandit_based_search} to the na\"ive algorithm for MIPS. In particular, Algorithm \ref{alg:bandit_based_search} is never worse in big-$O$ sample complexity than the na\"ive algorithm.

\label{subsec:gaps}
\textbf{Discussion of the importance of $\Delta$:}
In general, \algname takes only $O\left(\frac{1}{\Delta^2}\log \frac{n}{\delta \Delta}\right)$ computations per atom if there is reasonable heterogeneity among them.
As proven in Appendix 2 in \cite{bagaria2018medoids}, this is the case under a wide range of distributional assumptions on the $\mu_{i}$'s, e.g., when the $\mu_{i}$'s follow a sub-Gaussian distribution across the atoms.
These assumptions ensure that \algname has an overall complexity of $O\left( \frac{n}{\Delta^2} \log \frac{n}{\delta \Delta}\right)$ that is independent of $d$ when $d$ is sufficiently large and $\Delta$ does not depend on $d$.

At first glance, the assumption that each $\Delta_i$ (and therefore $\Delta$) does not depend on $d$ may seem restrictive. However, such an assumption actually applies under a reasonable number of data-generating models.
For example, if the atoms' coordinates are drawn from a latent variable model, i.e., the $\mu_i$'s are fixed in advance and the atoms' coordinates correspond to instantiations of a random variable with mean $\mu_i$, then $\Delta_i$ will be independent of $d$. 
As a concrete example, two users' $0/1$ ratings of movies may agree on 60\% of movies and their atoms' coordinates correspond to observations of a Bernoulli random variable with parameter $0.6$.
Other recent works provide further discussion on the conversion between an instance-wise bound like Equation \eqref{eqn:instance_bd} and an instance-independent bound that is independent of $d$ \cite{bagaria2018medoids,baharav2019ultra,tiwari2020banditpam,bagaria2021bandit,baharav2022approximate}.

%\iscomment{this subsection is great}

However, we note that in the worst case \algname may take $O(d)$ computations per atom when most atoms are equally good, for example in datasets where the atoms are symmetrically distributed around $\mathbf{q}$.
For example, if each atom's coordinates are drawn i.i.d. from the \textit{same} distribution, then the gaps $\Delta_i$ will scale inversely with $d$; we provide an example experiment on this type of dataset in Appendix \ref{app:symmetric}.

\label{subsec:weighted_sampling}
\textbf{Optimal weights for non-uniform sampling:}
Let $J \sim P_\mathbf{w}$ be a random variable following the categorical distribution $P_\mathbf{w}$, where $\mathbb{P}(J=j) = w_j\geq 0$ and $\sum_{j \in [d]} w_j = 1$. The arm parameter $\mu_i$ of an atom $i$ can be estimated by the unbiased estimator $X_{iJ} = \frac{1}{dw_J} v_{iJ} q_J$.
(Note that $d$ is fixed and known in advance). To see that $X_{iJ}$ is unbiased, we observe that $\mathbb{E}_{J \sim P_\mathbf{w}}[X_{iJ}] = \sum_{j \in [d]} w_j \frac{1}{dw_j} v_{ij} q_j = \sum_{j \in [d]} \frac{v_{ij} q_j}{d} = \mu_i$.

We are interested in finding the best weights $\mathbf{w}^*$, i.e., those that minimize the combined variance 
\begin{align} \label{eq:optimize}
    \argmin_{w_1,\ldots,w_d \geq 0} \sum_{i \in [n]} \text{Var}_{J \sim P_\mathbf{w}} [X_{iJ}], ~~~~s.t.~\sum_{j \in [d]} w_j = 1.
\end{align}

\begin{theorem}
\label{thm:optimal_weights}
The solution to Problem \eqref{eq:optimize} is
\begin{align}
\label{eqn:optimal_weights}
    w_j^* = \frac{\sqrt{q_j^2 \sum_{i \in [n]} v_{ij}^2}}{\sum_{j \in [d]} \sqrt{q_j^2 \sum_{i \in [n]} v_{ij}^2}},~~~~\text{for}~j=1,\ldots,d.
\end{align}
\end{theorem}

The proof of Theorem \ref{thm:optimal_weights} is provided in Appendix \ref{app:proofs}.

\textbf{Remark:}
In practice, computing the atom variance $\sum_{i \in [n]} v_{ij}^2$ requires $O(nd)$ operations and can be computationally prohibitive.
However, we may approximate $\sum_{i \in [n]} v_{ij}^2$ based on domain-specific assumptions. 
Specifically, if we assume that for each coordinate $j$, $q_j$ has a similar magnitude as $v_{ij}$'s, we can approximate $\frac{1}{n} \sum_{i \in [n]} v_{ij}^2 \approx q_j^2$ and set $w_j^* = \frac{q_j^2}{\sum_{j \in [d]} q_j^2}$. 
In the non-uniform sampling versions of \algnamenospace, we use an additional hyperparameter $\beta$ and let $w_j^* \propto q_j^{2\beta}$.
$\beta$ can be thought of as a temperature parameter which governs how uniformly (or not) we sample the coordinates based on the query vector's values.
We note that $\beta=1$ corresponds Equation \eqref{eqn:optimal_weights}.

The version we call \algnamenospace-$\alpha$ corresponds to taking the limit $\beta \rightarrow \infty$. In this case, we sort the query vector explicitly and sample coordinates in order of the sorted query vector; the sub-Gaussianity parameter used in \algnamenospace-$\alpha$ is then the same as that in the original problem with uniform sampling. While the sort incurs $O(d\text{log}d)$ cost, we find this still improves the overall sample complexity of the algorithm relative to the closest baseline when $O(d\text{log}d + n)$ is better than $O(n\sqrt{d})$, as is often the case in practice.

%% file: 7-experiments.tex
\section{Experiments}
\label{sec:experiments}
We empirically evaluate the performance of \algname and the non-uniform sampling version \algnamenospace-$\alpha$ on four synthetic and real-world datasets, comparing them to 8 state-of-art MIPS algorithms. 
We considered the two synthetic datasets, \texttt{NORMAL\_CUSTOM} and \texttt{CORRELATED\_NORMAL\_CUSTOM}, to assess the performance across a wide parameter range. 
We further considered the two real-world datasets, the Netflix Prize dataset ($n=6,000$, $d=400,000$) \cite{bennettNetflixPrize2007} and the Movie Lens dataset ($n=4,000$, $d=6,000$) \cite{movie2015}, to provide additional evaluations.
%\martin{Make sure results are shown for all 4 datasets.}
We compared our algorithms to 8 baseline MIPS algorithms: LSH-MIPS \cite{shrivastavaAsymmetricLSHALSH2014}, H2-ALSH-MIPS \cite{huang_accurate_2018}, NEQ-MIPS \cite{dai_norm-explicit_2020}, PCA-MIPS \cite{bachrachSpeedingXboxRecommender2014}, BoundedME \cite{liuBanditApproachMaximum2019}, Greedy-MIPS \cite{yuGreedyApproachBudgeted2017}, HNSW-MIPS \cite{Malkov2016a, morozovNonmetricSimilarityGraphs2018}. and NAPG-MIPS \cite{tan_norm_2021}.
%\martin{Make sure all XX baselines were included in all experiments.}
Throughout the experiments, we focus on the sample complexity, defined as the number of coordinate-wise multiplications performed.
Appendix \ref{app:datasets} provides additional details on our experimental settings.

\begin{center}
\begin{figure}[h]
\centering
\hspace{2em}
\begin{subfigure}{.35\textwidth}
    \centering
    \includegraphics[width=\linewidth]{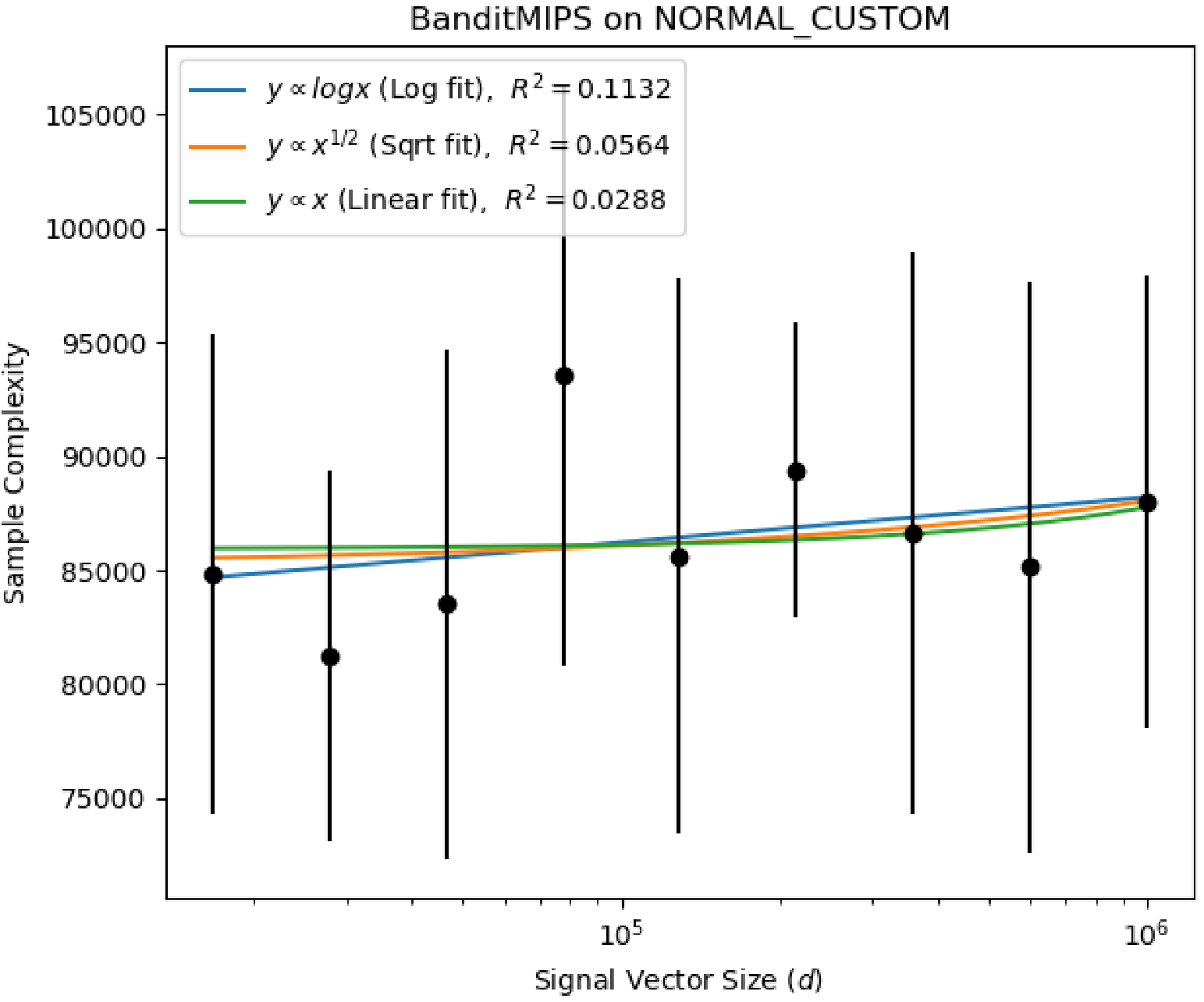}
    \caption{\label{scaling1:a}}
\end{subfigure}
\hspace{0.2em}
\begin{subfigure}{.39\textwidth}
    \centering
    \includegraphics[width=\linewidth]{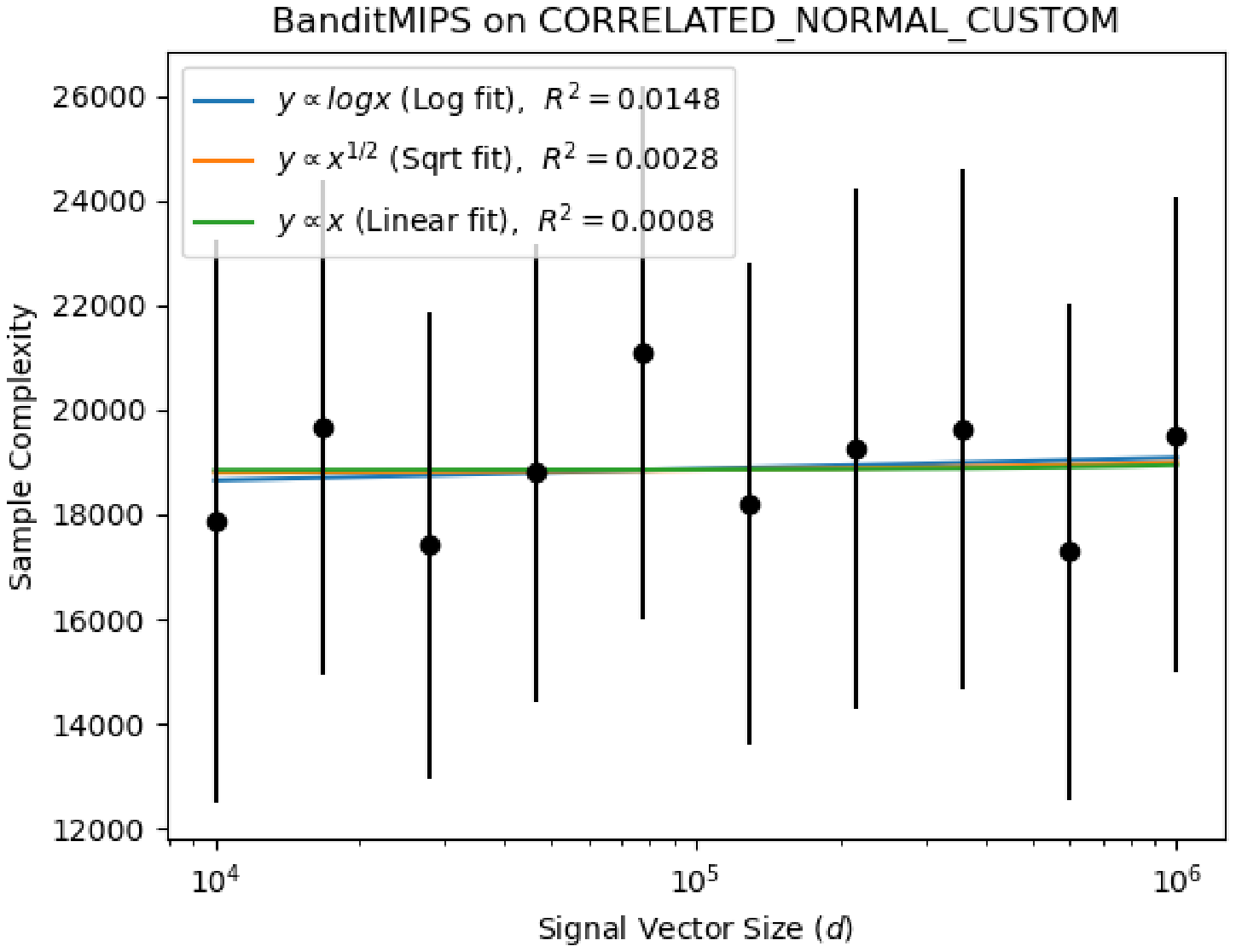}
    \caption{\label{scaling1:b}}
\end{subfigure}%
\newline
\begin{subfigure}{.37\textwidth}
    \centering
    \includegraphics[width=\linewidth]{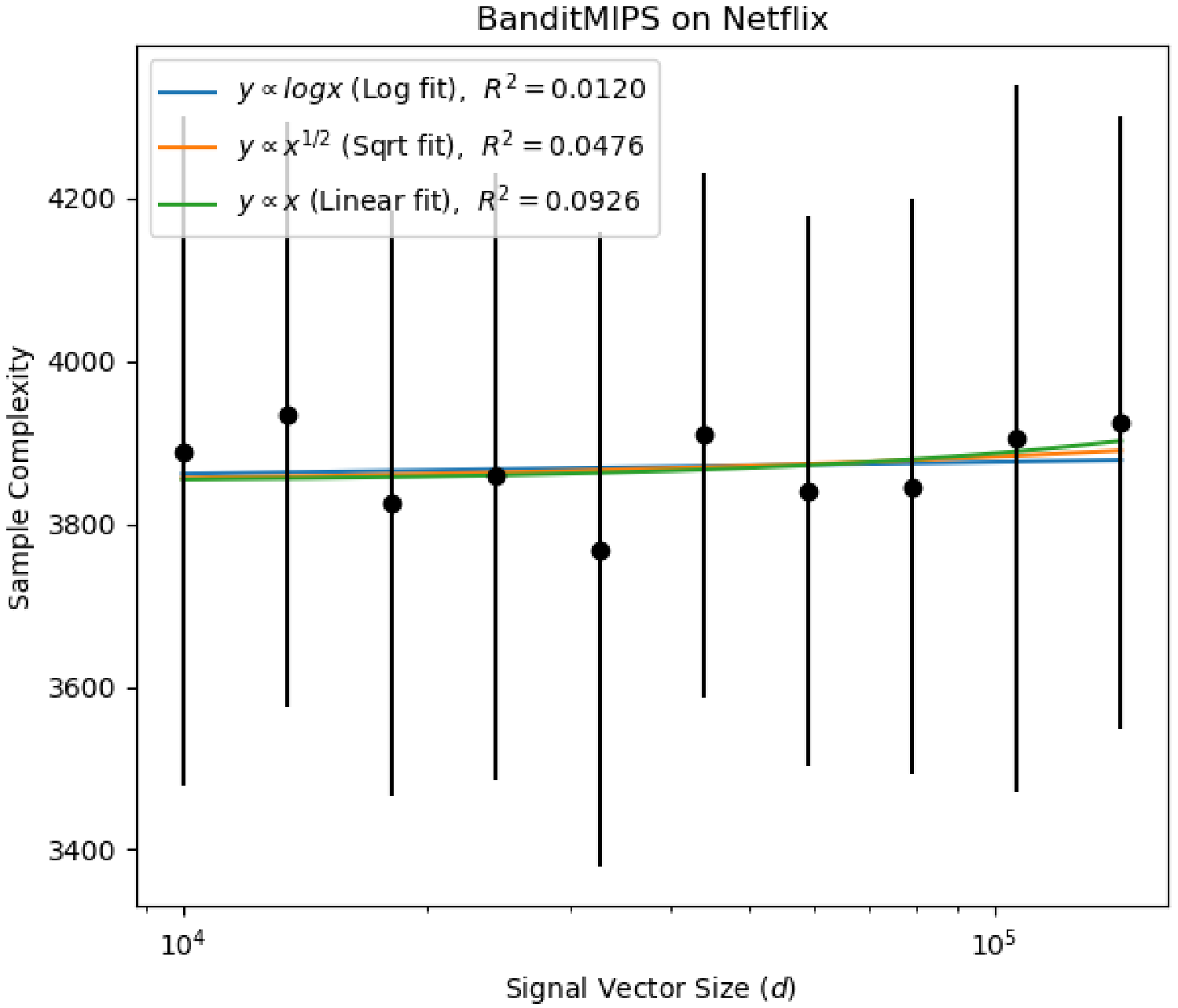}
    \caption{\label{scaling1:c}}
\end{subfigure}%
\begin{subfigure}{.37\textwidth}
    \centering
    \includegraphics[width=\linewidth]{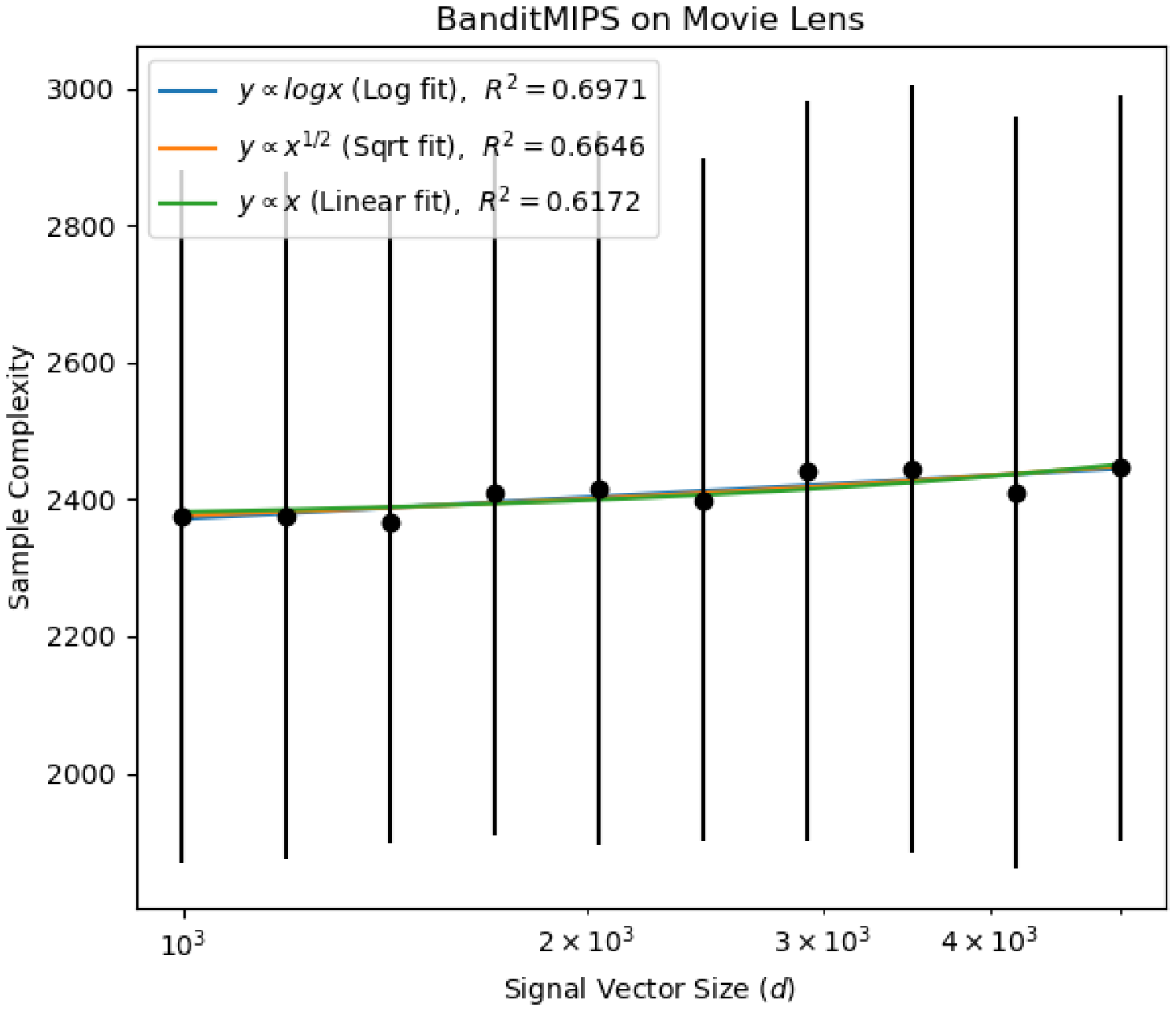}
    \caption{\label{scaling1:d}}
\end{subfigure}
\caption{Sample complexity of \algname for different values of $d$ on all four datasets. 95\% CIs are provided around the mean and are computed from 10 random trials. The sample complexity of \algname does not scale with $d$. Note that the values of $R^2$, the coefficient of determination, are similar for linear, logarithmic, and square root fits, which suggests the scaling is actually constant.}
\label{fig:bm_complexities1}
\end{figure}
\end{center}

\label{subsec:scaling}
\textbf{Scaling with $d$:}
We first assess the scaling with $d$ for \algname on the four datasets. We subsampled features from the full datasets, evaluating $d$ up to $1,000,000$ on simulated data and up to $400,000$ on real-world data. Results are reported in Figure \ref{fig:bm_complexities1}.
In all trials, \algname returns the correct answer to MIPS.
We determined that \algname did not scale with $d$ in all experiments, validating our theoretical results on the sample complexity. 

\begin{figure}
\centering
\hspace{2em}
\begin{subfigure}{.4\textwidth}
\centering
\includegraphics[width=\linewidth]{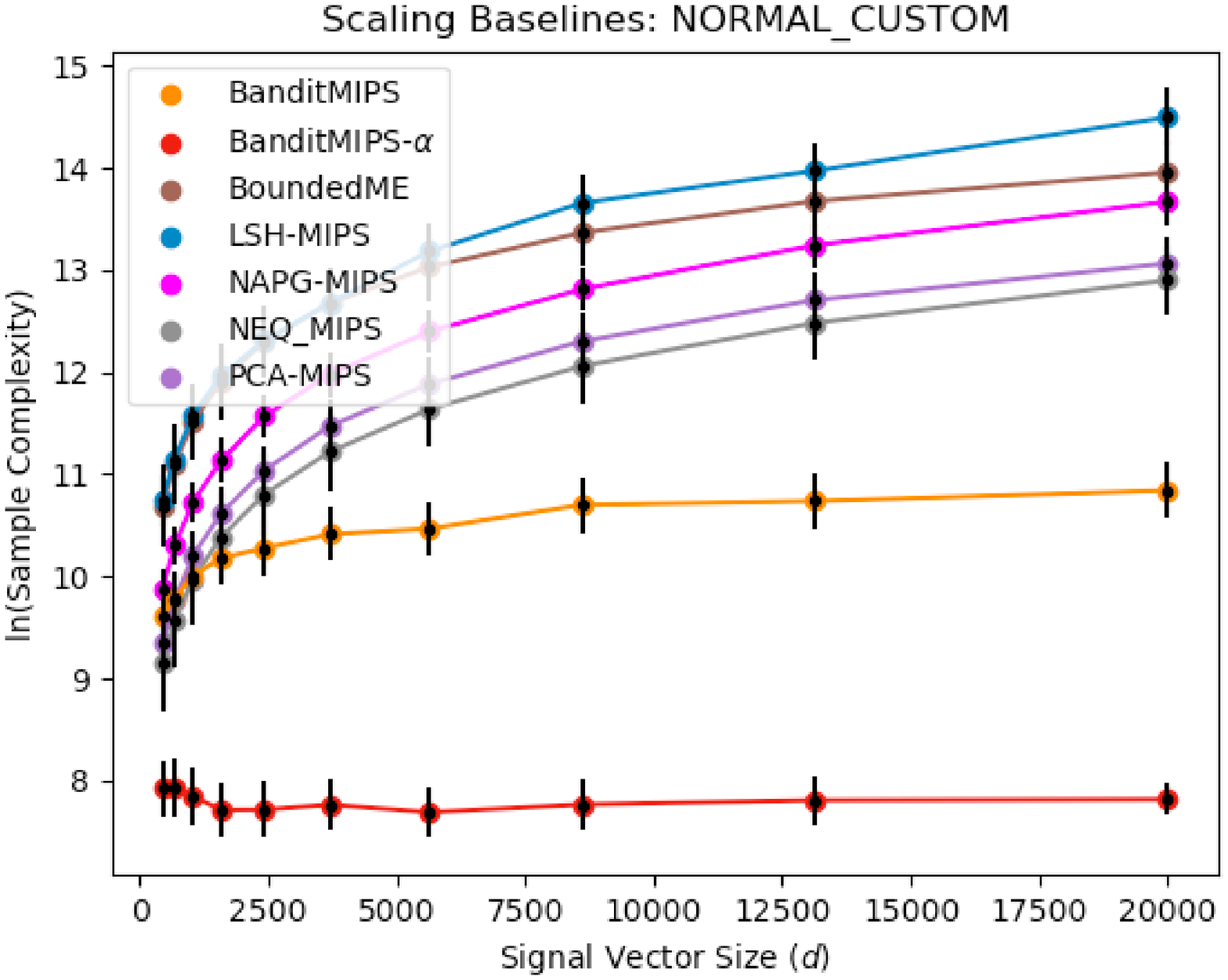}
\caption{\label{scaling2:a}}
\end{subfigure}%
\begin{subfigure}{.4\textwidth}
\centering
\includegraphics[width=\linewidth]{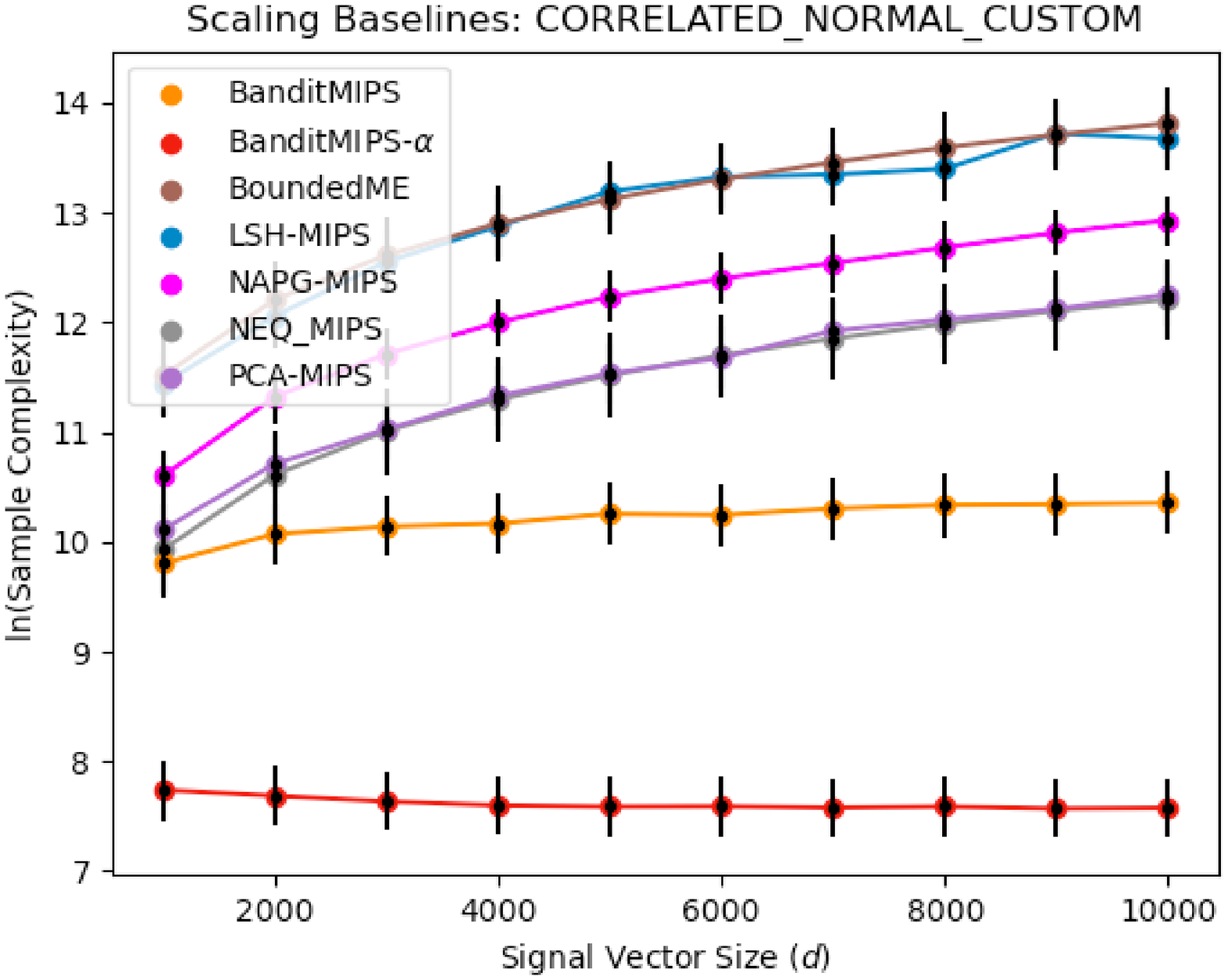}
\caption{\label{scaling2:b}}
\end{subfigure}
\newline
\begin{subfigure}{.4\textwidth}
\centering
\includegraphics[width=\linewidth]{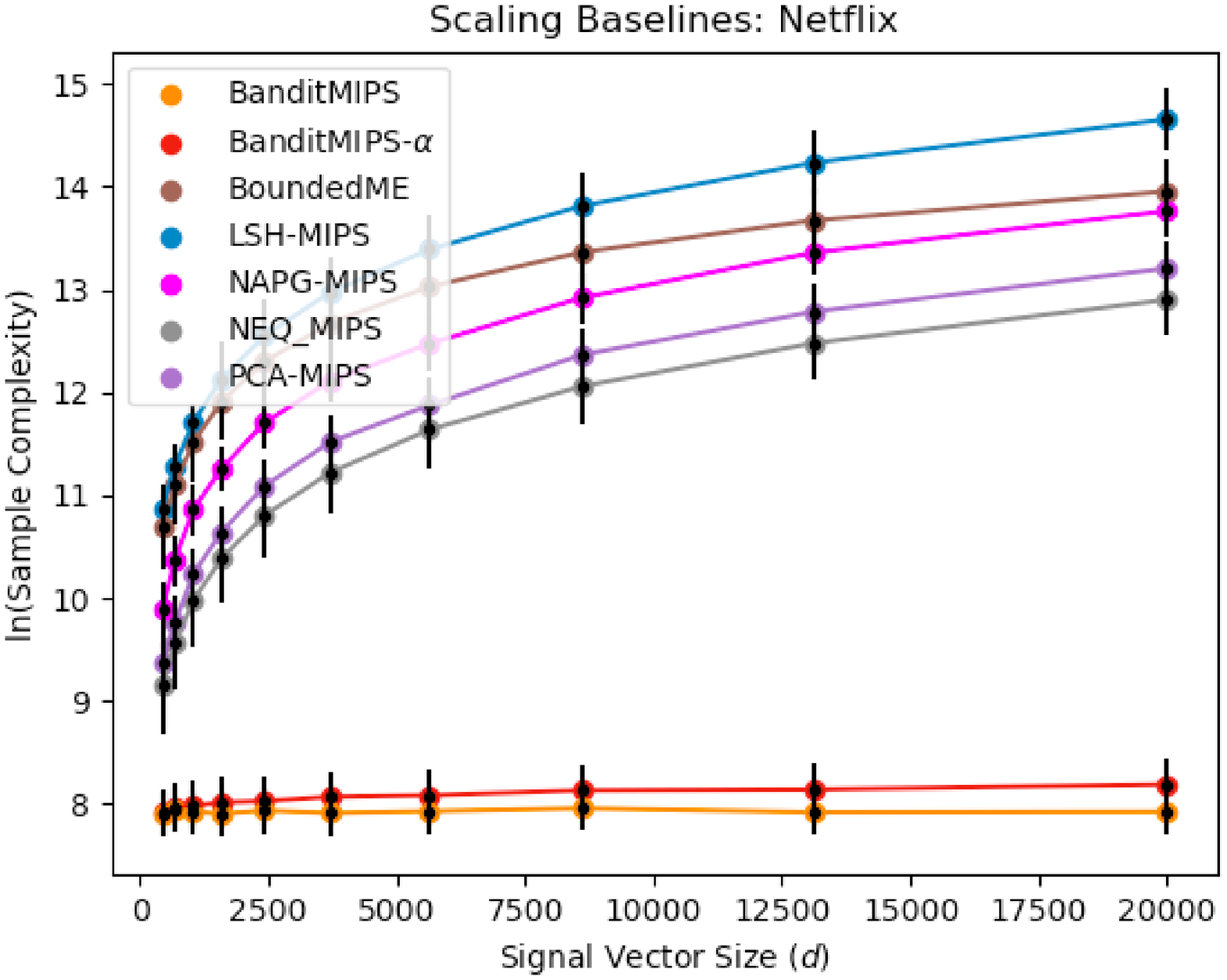}
\caption{\label{scaling2:c}}
\end{subfigure}%
\begin{subfigure}{.4\textwidth}
\centering
\includegraphics[width=\linewidth]{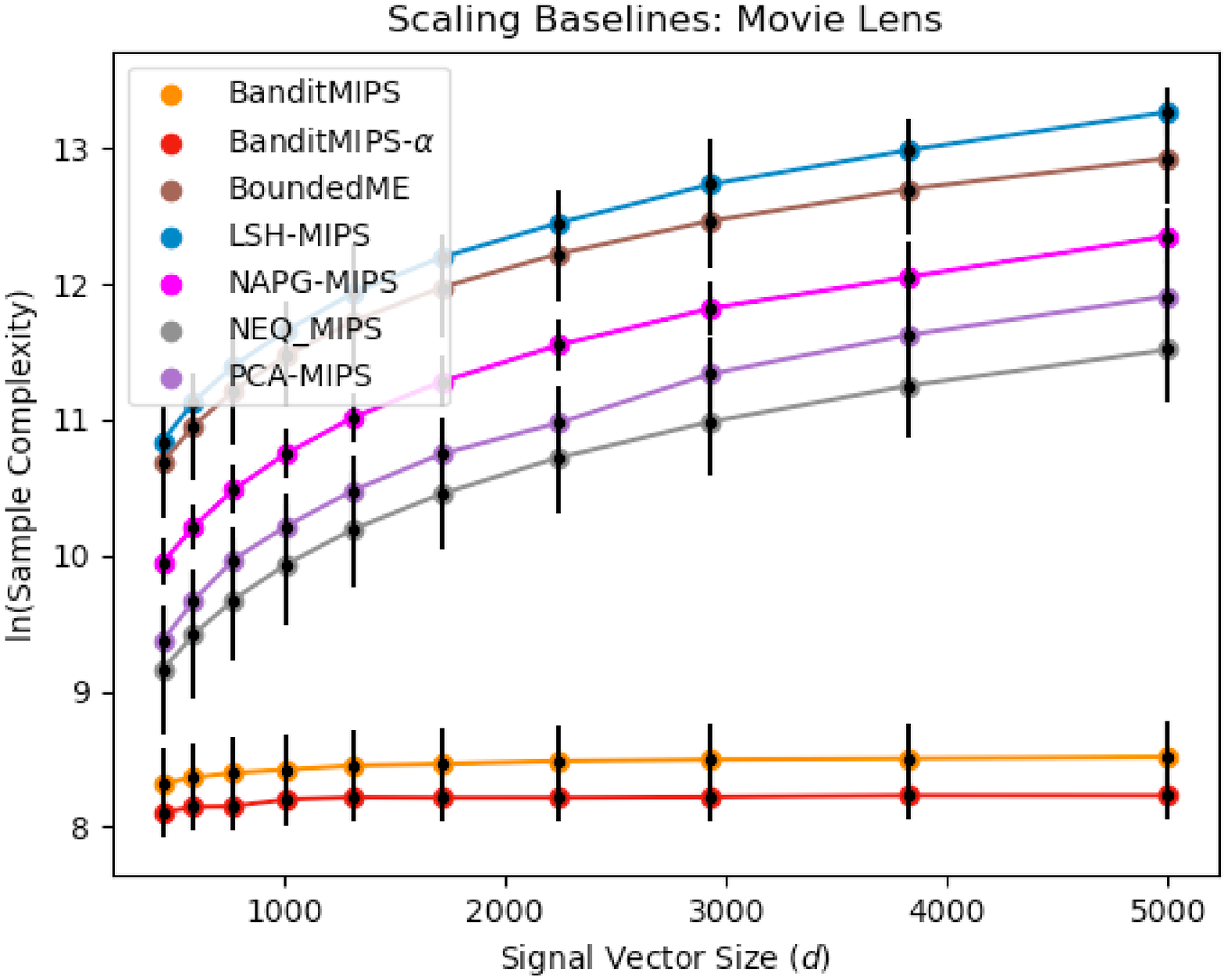}
\caption{\label{scaling2:d}}
\end{subfigure}
\caption{Comparison of sample complexity between \algnamenospace, \algnamenospace-$\alpha$, and other baseline algorithms for different values of $d$ across all four datasets. The $y$-axis is on a logarithmic scale. 95\% CIs are provided around the mean and are computed from 10 random trials. \algname and \algnamenospace-$\alpha$ outperformed other baselines.}
\label{fig:all_algo_complexities}
\end{figure}

\label{subsec:sample_complexities}
\textbf{Comparison of sample complexity:}
We next compare the sample complexity of \algname and \algnamenospace-$\alpha$ to 8 state-of-art MIPS algorithms on the four datasets across different values of $d$. 
We only used a subset of up to 20K features because some of the baseline algorithms were prohibitively slow for larger values of $d$.
Results are reported in Figure \ref{fig:all_algo_complexities}.
We omit GREEDY-MIPS from Figure \ref{fig:all_algo_complexities} because its sample complexity was significantly worse than all algorithms, and omit HNSW-MIPS as its performance was strictly worse than NAPG-MIPS (a related baseline).
In measuring sample complexity, we measure \textit{query-time} sample complexity, i.e., neglect the cost of preprocessing for the baseline algorithms; this is favorable to the baselines.
Nonetheless, our two algorithms substantially outperformed other algorithms on all four datasets, demonstrating their superiority in sample efficiency. 
For example, on the Movie Lens dataset, \algname and \algnamenospace-$\alpha$ are 20$\times$ and 27$\times$ faster than the closest baseline (NEQ-MIPS).
In addition, the non-uniform sampling version \algnamenospace-$\alpha$ outperformed the default version \algname in 3 out 4 datasets, suggesting the weighted sampling technique further improves sample efficiency. \algnamenospace-$\alpha$ demonstrated slightly worse performance than \algname on the Netflix dataset, possibly because the highest-value coordinates for the randomly sampled query vectors had low dot products with the atoms.

\label{subsec:prec1_tradeoff}
\textbf{Trade-off between speed and accuracy:}
Finally, we evaluate the trade-off between speed and accuracy by varying the error probability $\delta$ in our algorithm and the corresponding hyper-parameters in the baseline algorithms (see Appendix \ref{subsec:experimental_settings} for more details). 
As in \cite{liuBanditApproachMaximum2019}, we define the speedup of an algorithm to be: $\text{speedup} = \frac{\text{sample complexity of na\"ive algorithm}}{\text{sample complexity of compared algorithm}}. $
The accuracy is defined as the proportion of times each algorithm returns the true MIPS solution.
Results are reported in Figure \ref{fig:all_algo_complexities}. 
Our algorithms achieved the best tradeoff on all four datasets, again demonstrating the superiority of our algorithms in efficiently and accurately solving the MIPS problem. 
We also considered the $k$-MIPS setting where the goal was to find the top $k$ atoms. Results are reported for $k=5$ and $k=10$ in Appendix \ref{app:additional_experiments}. 
Our algorithms obtained a similar improvement over other baselines in these experiments.

\begin{figure}
\centering
\hspace{2em}
\begin{subfigure}{.4\textwidth}
\centering
\includegraphics[width=\linewidth]{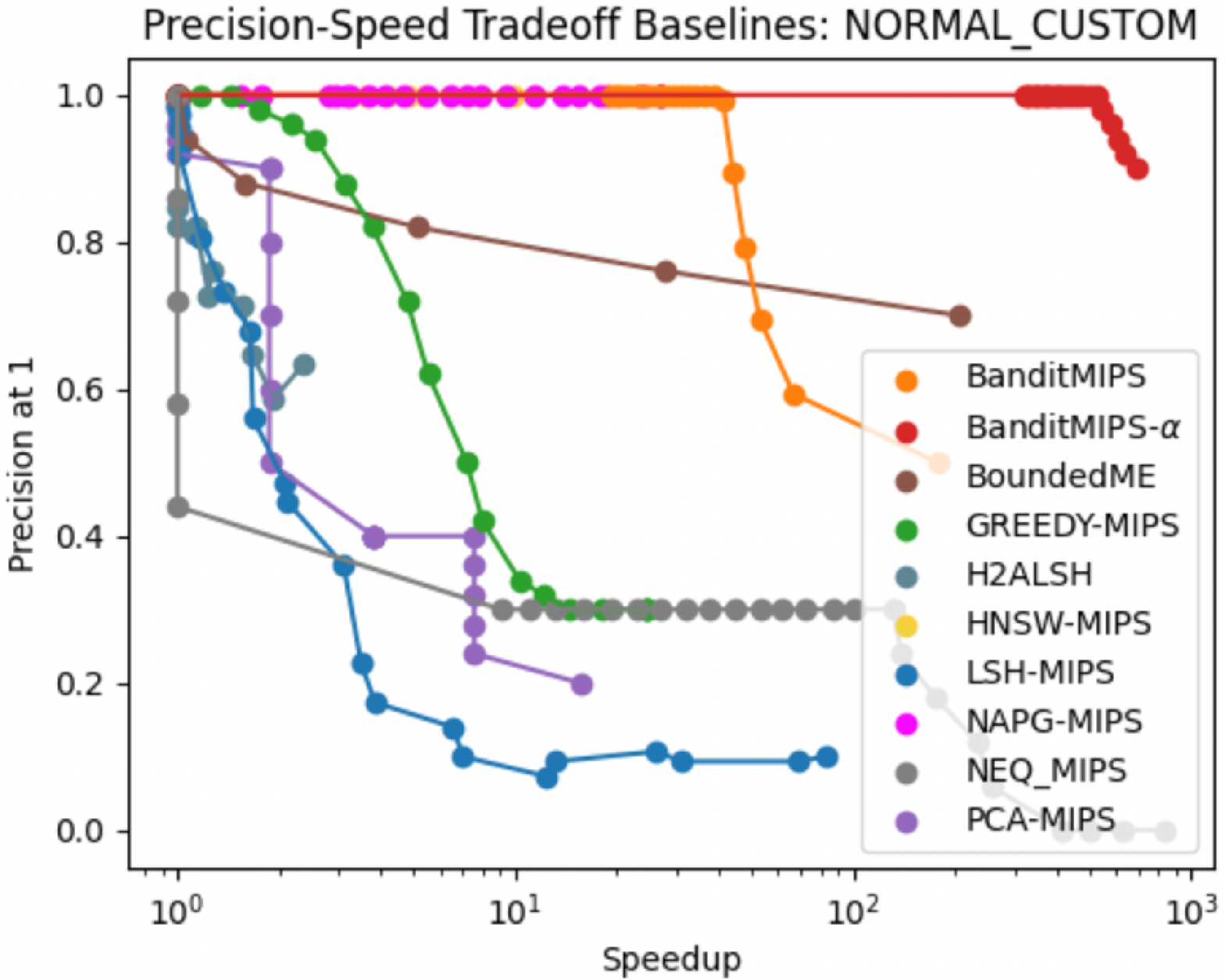}
\caption{\label{p1st:a}}
\end{subfigure}%
\begin{subfigure}{.4\textwidth}
\centering
\includegraphics[width=\linewidth]{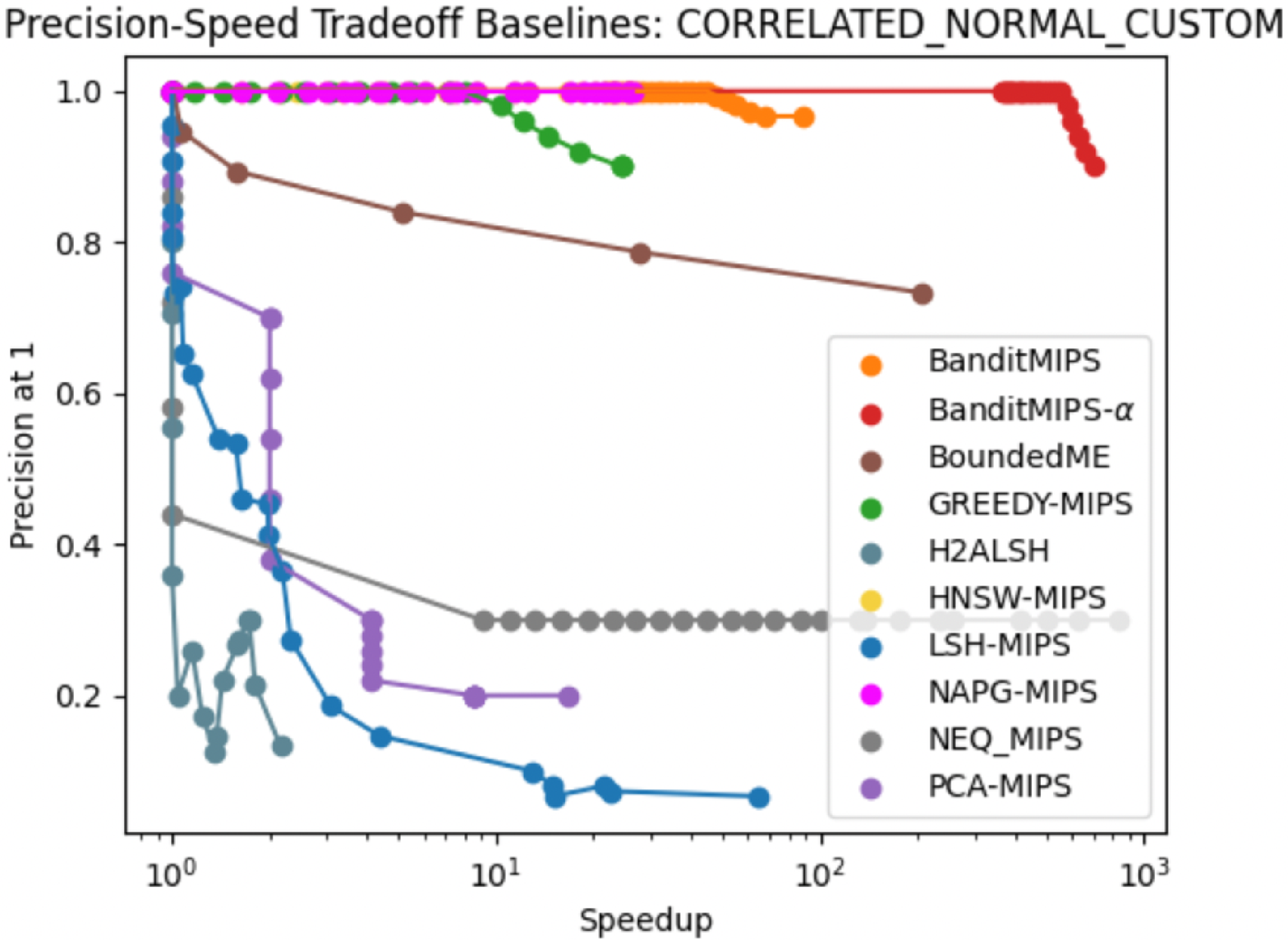}
\caption{\label{p1st:b}}
\end{subfigure}
\newline
\begin{subfigure}{.4\textwidth}
\centering
\includegraphics[width=\linewidth]{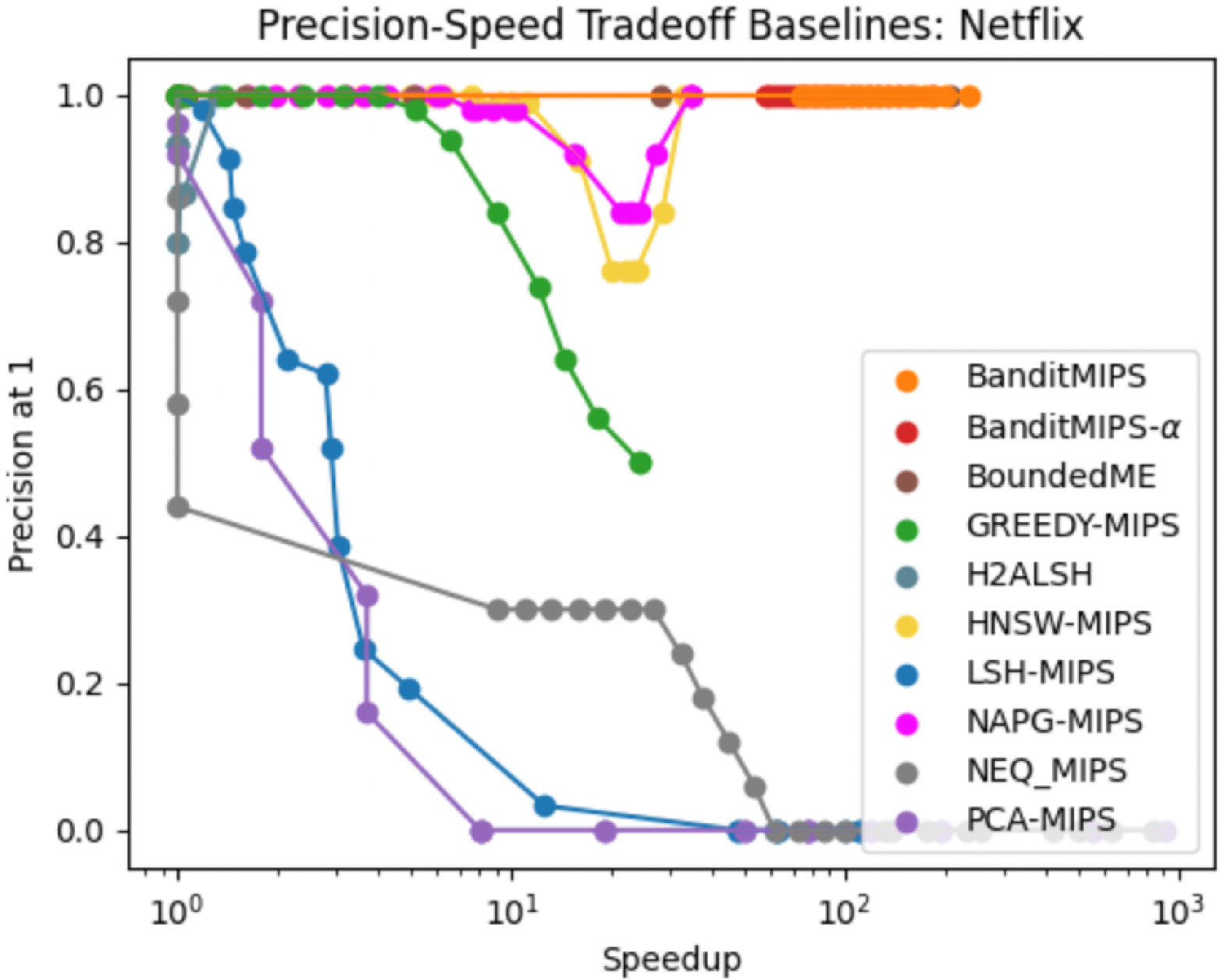}
\caption{\label{p1st:c}}
\end{subfigure}%
\begin{subfigure}{.4\textwidth}
\centering
\includegraphics[width=\linewidth]{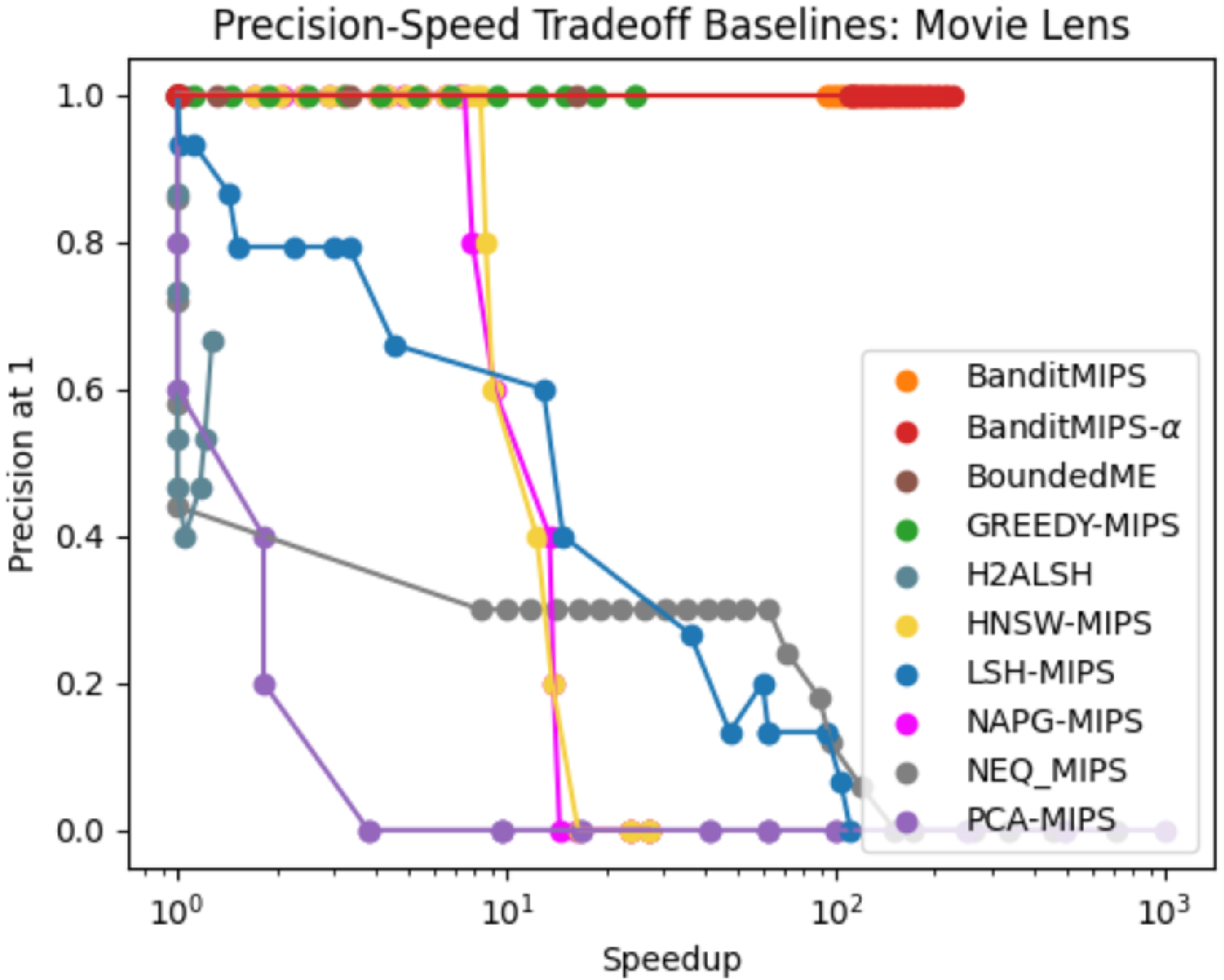}
\caption{\label{p1st:d}}
\end{subfigure}
\caption{Trade-off between accuracy (equivalent to precision@$1$) and speed for various algorithms across all four datasets. The $x$-axis represents the speedup relative to the naive $O(nd)$ algorithm and the $y$-axis shows the proportion of times an algorithm returned correct answer; higher is better. Each dot represents the mean across 10 random trials and the CIs are omitted for clarity. Our algorithms consistently achieve better accuracies at higher speedup values than the baselines.}
\label{fig:p1st1}
\end{figure}

\begin{figure}
\begin{subfigure}{.5\textwidth}
\centering
\includegraphics[width=\linewidth]{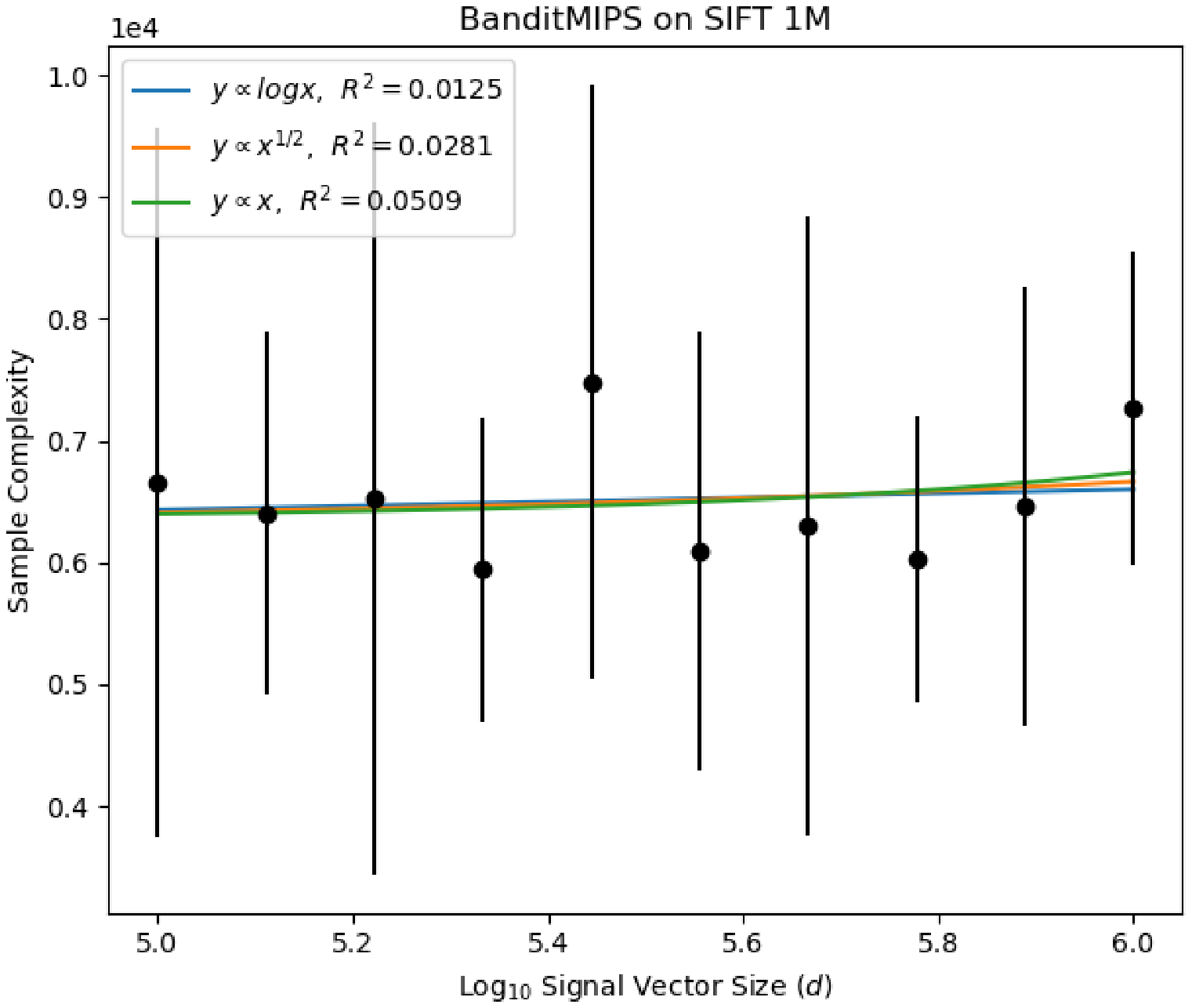}
\caption{}
\label{fig:bm_sift}
\end{subfigure}
\begin{subfigure}{.5\textwidth}
\centering
\includegraphics[width=\linewidth]{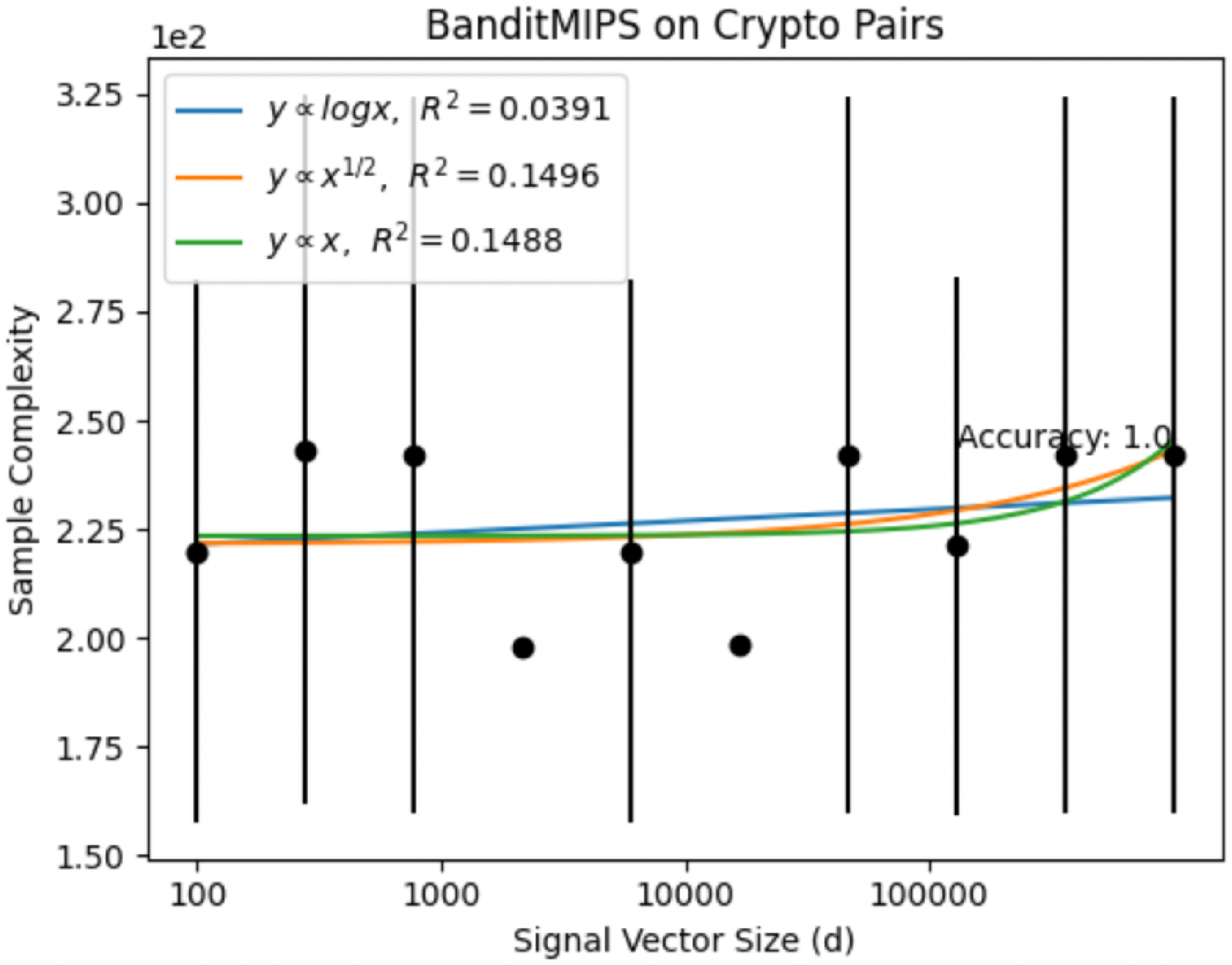}
\caption{}
\label{fig:bm_crypto}
\end{subfigure}
\caption{Sample complexity of \algname versus $d$ for the \texttt{Sift-1M} and \texttt{CryptoPairs} datasets. \algname scales as $O(1)$ with respect to $d$ for both datasets. Means and uncertainties were obtained by averaging over 5 random seeds. \algname returns the correct solution to MIPS in each trial.}
\label{fig:bm_scaling_sift_and_crypto}
\end{figure}

\label{subsec:real_world_high_dim}
\textbf{Real-world high-dimensional datasets:}
We also verify the $O(1)$ scaling with $d$ on two real-world, high-dimensional datasets: the \texttt{Sift-1M} \cite{jegou_product_2011} and \texttt{CryptoPairs} datasets \cite{400CryptoCurrency}.
The \texttt{Sift-1M} dataset consists of scale-invariant feature transform \cite{loweObjectRecognitionLocal1999} features of 128 different images; each image is an atom with $1$ million dimensions.
%In this setting, solving MIPS corresponds to finding the ``dimension'' of the original dataset that has the highest inner product with the query ``dimension''. 
The \texttt{CryptoPairs} dataset consists of the historical trading data of more than 400 trading pairs at 1 minute resolution reaching back until the year 2013.
On these datasets, \algname appears to scale as $O(1)$ with $d$ even to a million dimensions (Figure \ref{fig:bm_scaling_sift_and_crypto}). 
This suggests that the necessary assumptions outlined in Sections \ref{subsec:subgaussianity} and \ref{subsec:gaps} are satisfied on these real-world, high-dimensional datasets.
Note that the high dimensionality of these datasets makes them prohibitively expensive to run scaling experiments as in Section \ref{subsec:scaling} or the tradeoff experiments as in Section \ref{subsec:prec1_tradeoff} for baseline algorithms.

%% file: 8-conclusions_and_limitations.tex
% !TEX root = 0-main.tex

\section{Conclusions and Limitations}
\label{sec:conclusion}

In this work, we presented \algname and \algnamenospace-$\alpha$, novel algorithms for the MIPS problem.
In contrast with prior work, \algname requires no preprocessing of the data or that the data be nonnegative, and provides hyperparameters to trade off accuracy and runtime.
\algname scales better to high-dimensional datasets under reasonable assumptions and outperformed the prior state-of-the-art significantly.
%We expect our algorithms to demonstrate similar speedups in high-dimensional data when the assumptions in Section \ref{sec:theory} are satisfied.
Though the assumptions for \algname and \algnamenospace-$\alpha$ are often satisfied in practice, requiring them may be a limitation of our approach.
In particular, when many of the arm gaps are small, \algname will compute the inner products for the relevant atoms na\"ively. 
%Future work may focus on relaxing these assumptions or understanding how the performance of our algorithms degrades when they are violated.

%In Appendix \ref{app:preprocessing}, we also discuss combining \algname with preprocessing techniques to reduce its scaling with $n$ and its application to very-high-dimensional datasets.

%Finally, we note that it may be able to combine the approaches of \algname and \algnamenospace-$\alpha$ with other preprocessing techniques to reduce their scaling with $n$.
% \clearpage

%% file: 9.1-appendix_additional_related_work.tex
% !TEX root = 0-main.tex

\section{Additional Related Work}
\label{app:add_related_work}

In this appendix, we briefly describe other approaches attempt to reduce MIPS to a nearest neighbor search problem (NN). 
We note that the NN literature is extremely vast and has inspired the use of techniques based on permutation search \cite{naidanBN15}, inverted files \cite{AmatoS08}, vantage-point trees \cite{BoytsovN13b}, and more. 
The proliferation of NN algorithms has inspired several associated software packages \cite{Github:annoy, johnson2019billion, BoytsovN13} and tools for practical hyperparameter selection \cite{sun_automating_2023}.
However, MIPS is fundamentally different from and harder than NN because the inner product is not a proper metric function \cite{morozovNonmetricSimilarityGraphs2018}. 
Nonetheless, NN techniques have inspired many direct approaches to MIPS, including those that rely on $k$-dimensional or random projection trees \cite{dasgupta_random_2008}, concomitants of extreme order statistics \cite{CoCEOs, CoCEOs2, pham_sublinear_2020}, ordering permutations \cite{ChavezFN08}, principle component analysis (PCA) \cite{bachrachSpeedingXboxRecommender2014}, or hardware acceleration \cite{xiang_gaips_2021, abuzaid_index_2019}. 
All of these approaches require significant preprocessing that scales linearly in $d$, e.g., for computing the norms of the query or atom vectors, whereas \algname does not.

%% file: 9.2-appendix_proofs.tex
% !TEX root = 0-main.tex

\section{Proofs of Theorems}
\label{app:proofs}

In this appendix, we present the proofs of Theorems \ref{thm:specific} and \ref{thm:optimal_weights}.
%As above, for a given query $\mathbf{q}$, let $i^* = \argmax_{i \in [n]} \mu_{i}$ be the best atom and also let $\Delta_{i} \coloneqq \mu_{i^*} - \mu_{i}$ be the gap between other atoms and the best atom. 

\subsection{Proof of Theorem \ref{thm:specific}:}

% \martin{1. No need to reproduce the theorems, 2. One proof each subsection}
% \begin{theorem}
% \label{thm:specific_2}
% Assume $\exists~c_0 > 0$ s.t. $\forall~\delta'>0$, $d_\text{used}>0$, $C_{d_\text{used}}(\delta') < c_0\sqrt{\frac{\log n d^2_\text{used}/\delta'}{d_\text{used}}}$.
% % \iscomment{update?} 
% With probability at least $1-\delta$, \algname returns the correct solution to Equation \eqref{eqn:mips} and uses a total of $M$ computations, where
% \begin{align} 
% \label{eqn:instance_bd_2}
% M \leq \sum_{i \in [n]}  \min \left[ \frac{16c_0^2}{\Delta_{i}^2} \log \left( \frac{n}{\delta \Delta} \right) + 1, 2d \right]
% \end{align}
% \end{theorem}

\begin{proof}
Following the multi-armed bandit literature, we refer to each index $i$ as an arm and refer to its optimization object $\mu_{i}$ as the arm parameter. 
We sometimes abuse the terminology and refer to the atom $\mathbf{v}_i$ as the arm, with the meaning clear from context.
Pulling an arm corresponds to uniformly sampling a coordinate $J$ and evaluating $v_{iJ} q_J$ and incurs an $O(1)$ computation.
This allows us to focus on the number of arm pulls, which translates directly to coordinate-wise sample complexity.

%Our proof is similar to that in \cite{successiveelimination}. \martin{No necessary.}
First, we prove that with probability at least $1-\delta$, all confidence intervals computed throughout the algorithm are valid in that they contain the true parameter $\mu_{i}$'s.
For a fixed atom $\mathbf{v}_i$ and a given iteration of the algorithm, the $\left(1-\frac{\delta}{2 n d_\text{used}^2}\right)$ confidence interval satisfies 
\begin{align*}
    \Pr\left( \left| \mu_{i} - \hat \mu_{i} \right| > C_{d_\text{used}} \right) \leq 2e^{-C_{d_\text{used}}^2 d_\text{used} / 2 \sigma^2} \leq \frac{\delta}{2 n d_\text{used}^2}
\end{align*}

by Hoeffding's inequality and the choice of $C_{d_\text{used}} = \sigma \sqrt{\frac{2 \text{log}(4 n d^2_\text{used} / \delta)}{d_\text{used}+1}}$. 
%Note that there are at most $d$ rounds inside the \texttt{while} loop of \algname and hence at most  $n d$ confidence intervals computed across all arms and all steps of the algorithm.
%By a union bound over these $n d$ confidence intervals, we see that $\mu_{i} \in [\hat \mu_{i} - C_{i}, \hat \mu_{i} + C_{i}]$ for every arm $i$ and for every step of the algorithm with probability at least $1 - \frac{\delta}{d} > 1 - \delta$. 
For a fixed arm $i$, for any value of $d_\text{used}$ we have that the confidence interval is correct with probability at least $1 - \frac{\delta}{n}$, where we used the fact that $1 + \frac{1}{2^2} + \frac{1}{3^2} + \ldots = \frac{\pi^2}{6} < 2$.
%\martin{I don't follow this argument. Are you applying a union bound across rounds and using the fact that $1 + 1/2^2 + 1/3^2 + ... = \pi^2/6$? If so, be explicit.}
By another union bound over all $n$ arm indices, all confidence intervals constructed by the algorithm are correct with probability at least $1 - \delta$.

Next, we prove the correctness of  \algnamenospace.
Let $i^* = \argmax_{i \in [n]} \mu_{i}$ be the desired output of the algorithm.
First, observe that the main \texttt{while} loop in the algorithm can only run $d$ times, so the algorithm must terminate.
Furthermore, if all confidence intervals throughout the algorithm are valid, which is the case with probability at least $1-\delta$, $i^*$ cannot be removed from the set of candidate arms. 
Hence, $\mathbf{v}_{i^*}$ (or some $\mathbf{v}_i$ with $\mu_{i} = \mu_{i^*}$) must be returned upon termination with probability at least $1-\delta$. This proves the correctness of Algorithm \ref{alg:bandit_based_search}.

Finally, we examine the complexity of \algnamenospace. 
Let $d_{\text{used}}$ be the total number of arm pulls computed for each of the arms remaining in the set of candidate arms at a given iteration in the algorithm.
Note that for any suboptimal arm $i \ne i^*$ that has not left the set of candidate arms $\mathcal{S}_{\text{solution}}$, we must have
$C_{d_\text{used}} \leq c_0 \sqrt{ \frac{\log (1/\delta)}{d_{\text{used}}}}$ by assumption (and this holds for our specific choice of $C_{d_\text{used}}$ in Algorithm \ref{alg:bandit_based_search}).
With $\Delta_{i} = \mu_{i^*} - \mu_{i}$, if $d_{\text{used}} > \frac{16c_0^2}{\Delta_{i}^2} \log\frac{n}{\delta \Delta_i}$, then
\begin{align*}
4C_{d_\text{used}} &\leq 4 c_0 \sqrt{  \frac{{ \log\frac{n}{\delta \Delta_i} }}{{d_{\text{used}}}  }} < \Delta_{i}
\end{align*}
%\martin{Should the numerator in the 2nd term be $\log (nd^2_\text{used}/\delta)$ instead? Good to include more intermediate steps.}
% \iscomment{should it be 4 instead of 2 above?}
Furthermore, 
\begin{align*}
    \hat \mu_{i^*} - C_{d_\text{used}} &\geq \mu_{i^*} - 2C_{d_\text{used}} \\
    &= \mu_{i} + \Delta_{i} - 2C_{d_\text{used}} \\
    %&\overset{ \Delta_{i} > 2(C_{i} + C_{i^*})}{>}
    &> \mu_{i} + 2 C_{d_\text{used}} \\
    &> \hat \mu_{i} + C_{d_\text{used}}
\end{align*}
which means that $i$ must be removed from the set of candidate arms by the end of that iteration.

Hence, the number of data point computations $M_{i}$ required for any arm $i \ne i^*$ is at most
\begin{align*}
M_{i} \leq \min \left[ \frac{16c_0^2}{\Delta_{i}^2} \log\frac{n}{\delta \Delta_i} + 1, 2d \right]
\end{align*}

where we used the fact that the maximum number of computations for any arm is $2d$ when sampling with replacement.
Note that bound this holds simultaneously for all arms $i$ with probability at least $1-\delta$.
% We conclude that the total number of arm pulls $M$ satisfies
% \begin{align*}
% \mathbb{E}[M] & \leq \mathbb{E}[M | \text{ all confidence intervals are correct}] + \delta (2nd) \\
% & \leq \sum_{i \in [n]}  \min \left[ \frac{16c_0^2}{\Delta_{i}^2} \log\frac{nd^2}{\delta} + 1, 2d \right] + 2 \delta n d
% \end{align*}
We conclude that the total number of arm pulls $M$ satisfies
\begin{align*}
M & \leq \sum_{i \in [n]}  \min \left[ \frac{16c_0^2}{\Delta_{i}^2} \log\frac{n}{\delta \Delta_i} + 1, 2d \right]
\end{align*}
with probability at least $1-\delta$. 

%\martin{Can you add an argument to provide the upper bound for $\mathbb{E}[M]$? Let $E$ be the event that all CIs hold. Then $\mathbb{E}[M] = \mathbb{E}[M | E] P(E) + \mathbb{E}[M | \bar{E}] (1-P(E))$. However, in this case, your probability of $P(\bar{E})$ may not be small enough ... Let's discuss.}

As argued before, since each arm pull involves an $O(1)$ computation, $M$ also corresponds to the total number of operations up to a constant factor.
\end{proof}

%%%%%%%%%%%%%%%%%%%%%%%%%%%%%%%%%%%
%%%%%%%%%%%%%%%%%%%%%%%%%%%%%%%%%%%%%%%%%%%%%%%%%%%%%%%%%%%%%%%%%%%%%%

% \begin{theorem}
% \label{thm:optimal_weights_2}
% The solution to Problem \eqref{eq:optimize} is
% \begin{align}
%     w_j^* = \frac{\sqrt{q_j^2 \sum_{i \in [n]} v_{ij}^2}}{\sum_{j \in [d]} \sqrt{q_j^2 \sum_{i \in [n]} v_{ij}^2}},~~~~\text{for}~j=1,\ldots,d.
% \end{align}
% \end{theorem}

\subsection{Proof of Theorem \ref{thm:optimal_weights}}

\begin{proof}
Since all the $X_{iJ}$'s are unbiased, optimizing Problem \eqref{eq:optimize} is equivalent to minimizing the combined second moment
\begin{align}
    \sum_{i \in [n]} \mathbb{E}_{J \sim P_\mathbf{w}} [X_{iJ}^2] & = \sum_{i \in [n]} \sum_{j \in [d]} \frac{1}{d^2 w_j} q_j^2 v_{ij}^2 \\
    & = \sum_{j \in [d]} \left( \frac{1}{d^2 w_j} q_j^2 \sum_{i \in [n]} v_{ij}^2 \right). 
\end{align}
The Lagrangian is given by
\begin{align}
    \mathcal{L}(\mathbf{w}, \nu) = \sum_{j \in [d]} \left( \frac{1}{d^2 w_j} q_j^2 \sum_{i \in [n]} v_{ij}^2 \right) + \nu 
    \left( 1 - \sum_{j \in [d]} w_j \right).
\end{align}
Furthermore, the derivatives are
\begin{align}
    & \frac{\partial \mathcal{L}(\mathbf{w}, \nu)}{\partial w_j} = - \frac{q_j^2 \sum_{i \in [n]} v_{ij}^2}{d^2 w_j^2} - \nu\\
    & \frac{\partial \mathcal{L}(\mathbf{w}, \nu)}{\partial \mu} = 1 - \sum_{j \in [d]} w_j.
\end{align}

By the Karush-Kuhn-Tucker (KKT) conditions, setting the derivatives to 0 gives 
\begin{align}
    w_j^* = \frac{\sqrt{q_j^2 \sum_{i \in [n]} v_{ij}^2}}{\sum_{j \in [d]} \sqrt{q_j^2 \sum_{i \in [n]} v_{ij}^2}}~~~~\text{ for }~j=1,\ldots,d.
\end{align}
\end{proof}

%% file: 9.3-appendix_datasets.tex
% !TEX root = 0-main.tex

\section{Description of Datasets}
\label{app:datasets}

Here, we provide a more detailed description of the datasets used in our experiments. \\

\subsection{Synthetic Datasets}

In the \texttt{NORMAL\_CUSTOM} dataset, a parameter $\theta_i$ is drawn for each atom from a standard normal distribution, then each coordinate for that atom is drawn from $\mathcal{N}(\theta_i, 1)$. The signals are generated similarly.

In the \texttt{CORRELATED\_NORMAL\_CUSTOM} dataset, a parameter $\theta$ is for the signal $\mathbf{q}$ from a standard normal distribution, then each coordinate for that signal is drawn from $\mathcal{N}(\theta, 1)$. Atom $\mathbf{v}_i$ is generated by first sampling a random weight $w_i \sim \mathcal{N}(0, 1)$; then atom $\mathbf{v}_i$ is set to $w_i \mathbf{q}$ plus Gaussian noise.

Note that for the synthetic datasets, we can vary $n$ and $d$. The values of $n$ and $d$ chosen for each experiment are described in Subsection \ref{subsec:experimental_settings}.

\subsection{Real-world datasets}

\textbf{Netflix Dataset:} We use a subset of the data from the Netflix Prize dataset \cite{bennettNetflixPrize2007} that contains the ratings of 6,000 movies by 400,000 customers.
%We perform standard preprocessing to remove movies with very few ratings and impute missing values \cite{bennettNetflixPrize2007}.
%Its subset records the ratings from 400k customers on 6k movies. 
%After cleaning up data and filling missing values by fitting Collaborative Filtering model to the data and predicting them, we are able to get the matrix of size (1350, 143458). 
%Finally, we perform an SVD factorization on the dataset (with the number of factors corresponding to the data dimensionality) to obtain the user and movie vectors.
We impute missing ratings by approximating the data matrix via a low-rank approximation. Specifically, we approximate the data matrix via a 100-factor SVD decomposition. 
The movie vectors are used as the query vectors and atoms and $d$ corresponds to the number of subsampled users. \\
%Each row represents the ratings on each movie given by 143458 customers.

\textbf{Movie Lens Dataset:} We use Movie Lens-1M dataset \cite{movie2015}, which consists of 1 million ratings of 4,000 movies by 6,000 users. As for the Netflix dataset, we impute missing ratings by obtaining a low-rank approximation to the data matrix. Specifically, we perform apply a Non-negative Matrix Factorization (NMF) with 15 factors to the dataset to impute missing values.
The movie vectors are used as the query vectors and atoms, with $d$ corresponding to the number of subsampled users. \\

We note that for all  datasets, the coordinate-wise inner products are sub-Gaussian random variables. In particular, this means the assumptions of Theorem \ref{thm:specific} are satisfied and we can construct confidence intervals that scale as $O\left(\sqrt{\frac{\log 1/\delta'}{d'}}\right)$. We describe the setting for the sub-Gaussianity parameters in Section \ref{subsec:experimental_settings}.

\subsection{Experimental Settings}
\label{subsec:experimental_settings}

\textbf{Scaling Experiments:} In all scaling experiments, $\delta$ and $\epsilon$ were both set to $0.001$ for \algname and \algnamenospace-$\alpha$. For the \texttt{NORMAL\_CUSTOM} and \texttt{CORRELATED\_NORMAL\_CUSTOM} datasets, the sub-Gaussianity parameter was set to $1$. For the Netflix and Movie Lens datasets, the sub-Gaussianity parameter was set to $25$. For the \texttt{CryptoPairs}, \texttt{SIFT-1M}, and \texttt{SimpleSong} datasets described in Appendix \ref{app:high_dim}, the sub-Gaussianity parameters were set to $2.5e9$, $6.25e5$, and $25$, respectively. The number of atoms was set to $100$ and all other atoms used default values of hyperparameters for their sub-Gaussianity parameters.

\textbf{Tradeoff Experiments:} For the tradeoff experiments, the number of dimensions was fixed to $d = 10,000$.
The various values of speedups were obtained by varying the hyperparameters of each algorithm.
For NAPG-MIPS and HNSW-MIPS, for example, $M$ was varied from 4 to 32, $ef\_constructions$ was varied from $2$ to $500$, and $ef\_searches$ was varied from $2$ to $500$.
For Greedy-MIPS, $budget$ varied from 2 to 999. For LSH-MIPS, the number of hash functions and hash values vary from 1 to 10.
For H2ALSH, $\delta$ varies from $\frac{1}{2^4}$ to $\frac{1}{2}$, $c_0$ varies from 1.2 to 5, and $c$ varies from 0.9 to 2. 
For NEQ-MIPS, the number of codewords and codebooks vary from 1 to 100. 
For \algname, \algnamenospace-$\alpha$, and BoundedME, speedups were obtained by varying $\delta$ from $\frac{1}{10^{10}}$ to  $0.99$ and $\epsilon$ from $\frac{1}{10^{10}}$ to $3$. In our code submission, we include a one-line script to reproduce all of our results and plots.

All experiments were run on a 2019 Macbook Pro with a 2.4 GHz 8-Core Intel Core i9 CPU, 64 GB 2667 MHz DDR4 RAM, and an Intel UHD Graphics 630 1536 MB graphics card. Our results, however, should not be sensitive to hardware, as we used hardware-independent performance metrics (the number of coordinate-wise multiplications) for our results. 

%% file: 9.4-appendix_additional_experiments.tex
% !TEX root = 0-main.tex

\section{Additional Experimental Results}
\label{app:additional_experiments}

\begin{center}
\begin{figure}
\begin{subfigure}{.5\textwidth}
\centering
\includegraphics[width=\linewidth]{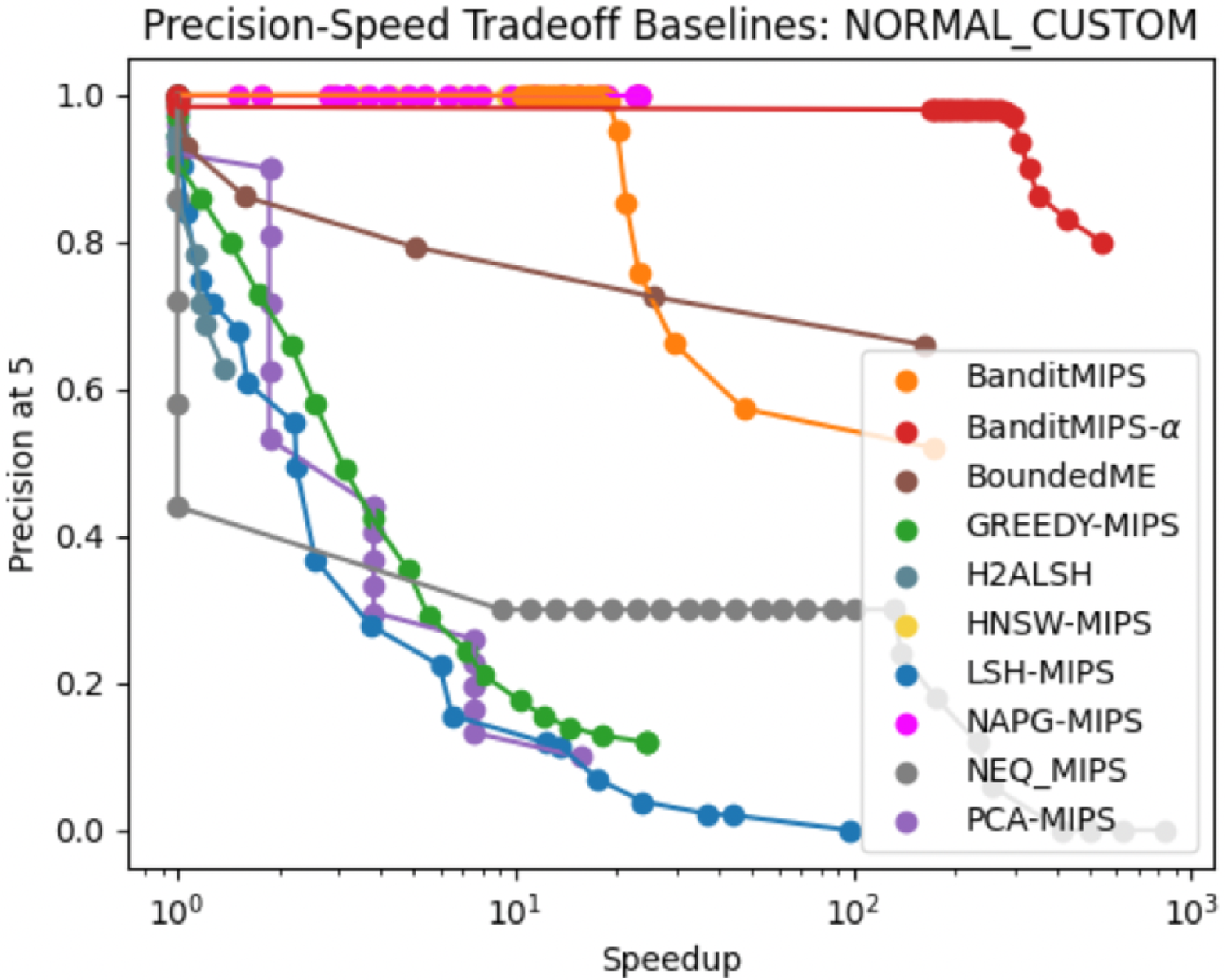}
\caption{}
\label{p5st:a}
\end{subfigure}
\begin{subfigure}{.5\textwidth}
\centering
\includegraphics[width=\linewidth]{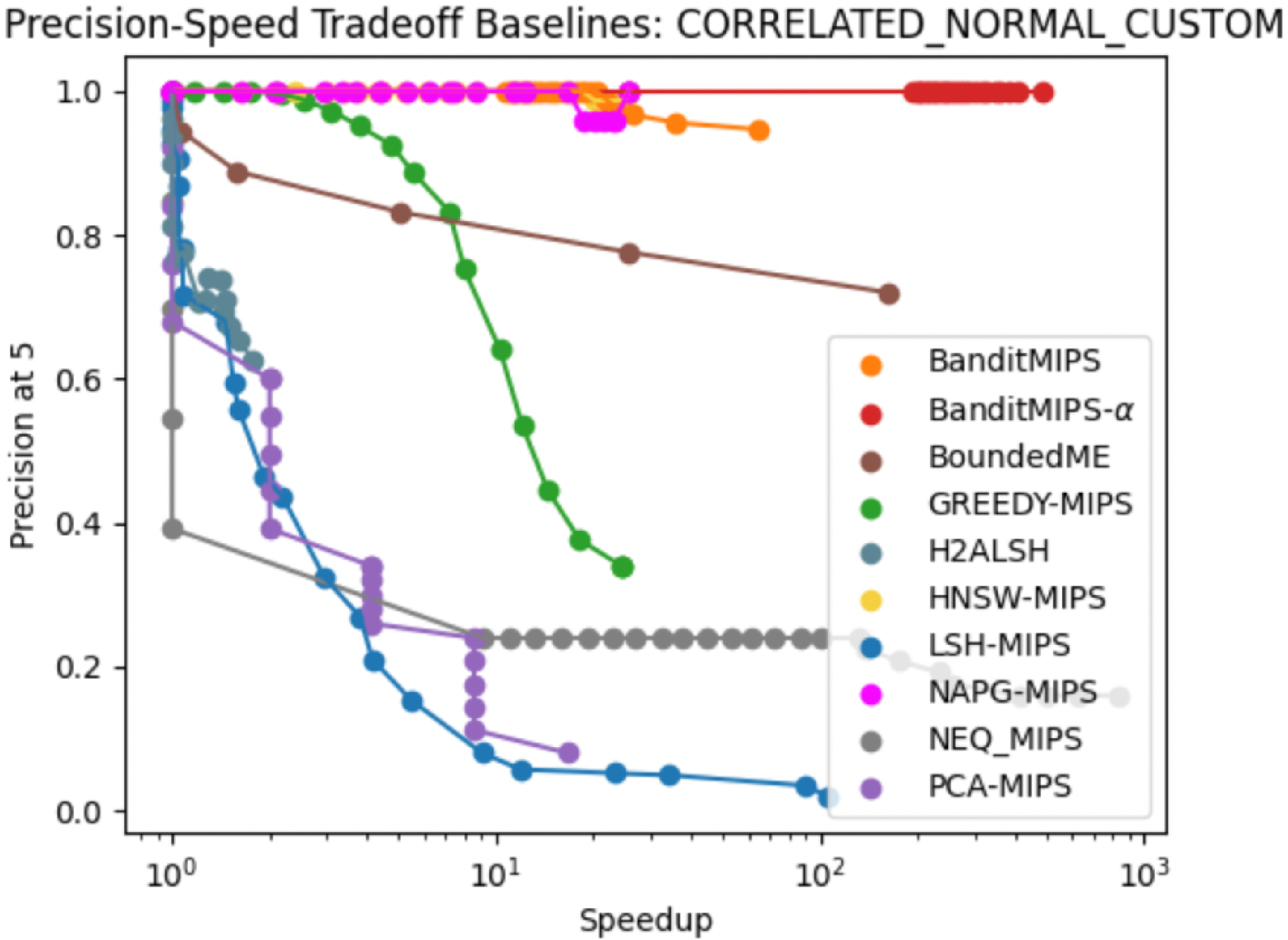}
\caption{}
\label{p5st:b}
\end{subfigure}
\begin{subfigure}{.5\textwidth}
\centering
\includegraphics[width=\linewidth]{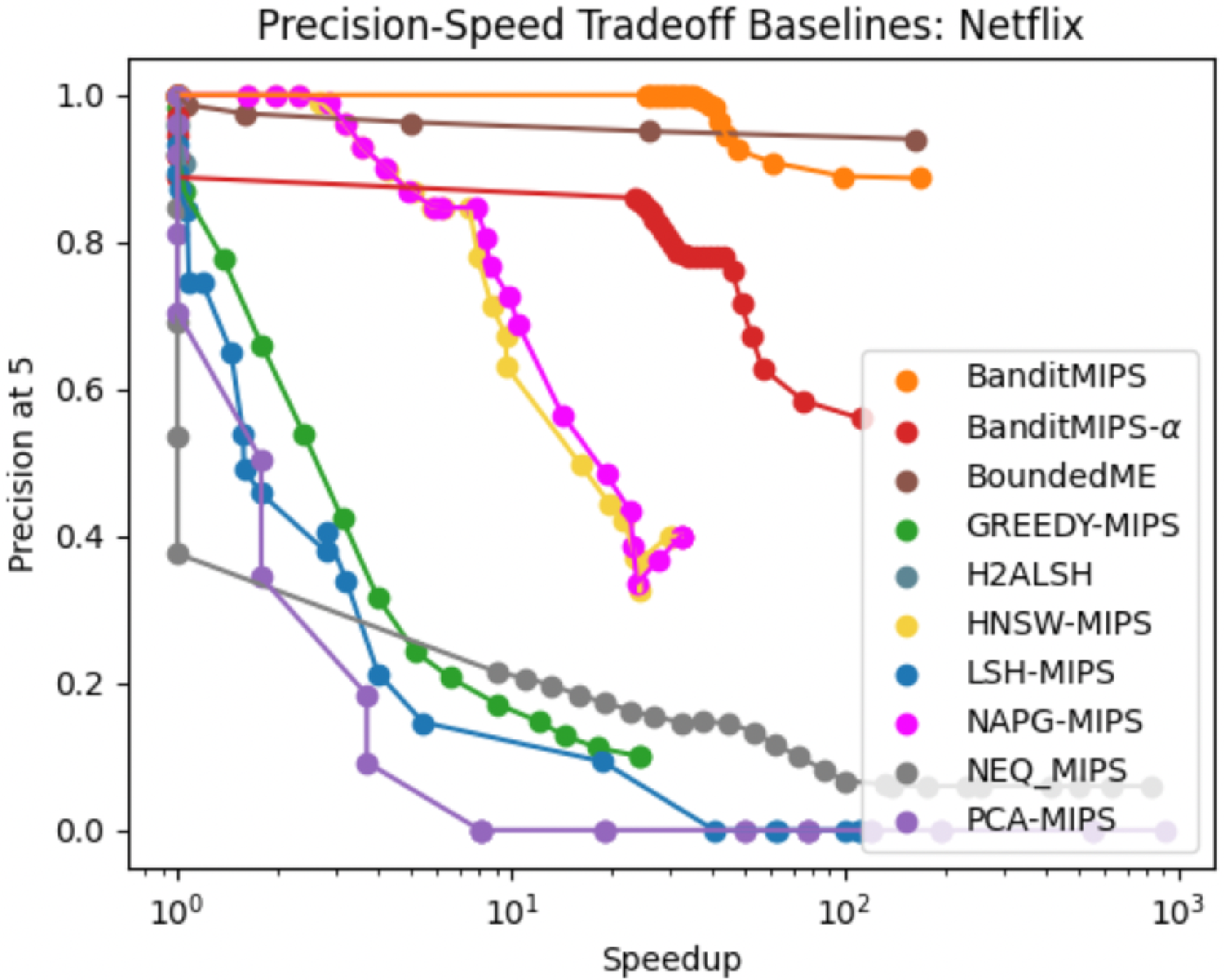}
\caption{}
\label{p5st:c}
\end{subfigure}
\begin{subfigure}{.5\textwidth}
\centering
\includegraphics[width=\linewidth]{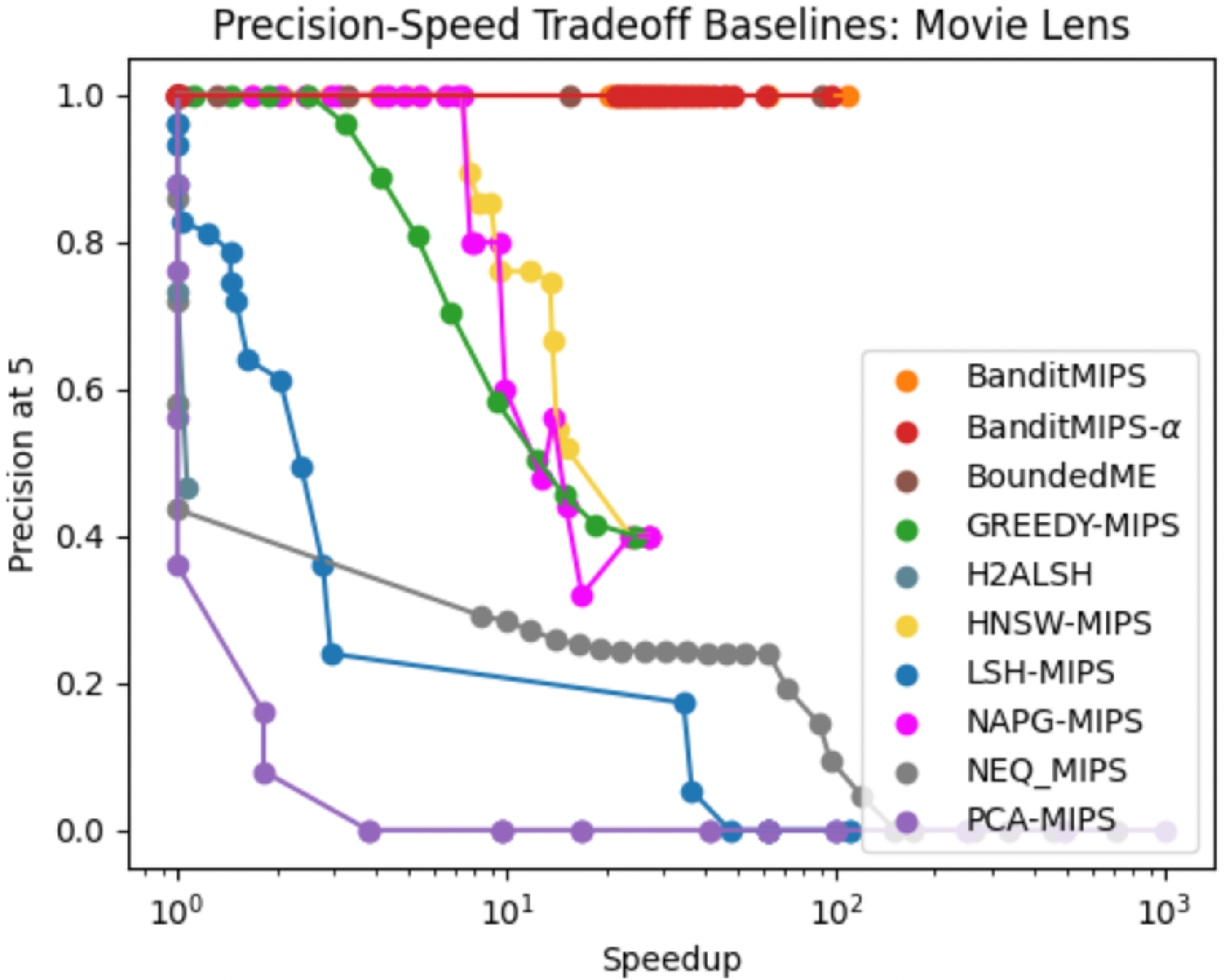}
\caption{}
\label{p5st:d}
\end{subfigure}
\caption{Trade-off between top-$5$ accuracy (precision@$5$) and speed for various algorithms across all four datasets. The $x$-axis represents the speedup relative to the naive $O(nd)$ algorithm and the $y$-axis shows the proportion of times an algorithm returned correct answer; higher is better. Each dot represents the mean across 10 random trials and the CIs are omitted for clarity. Our algorithms consistently achieve better accuracies at higher speedup values than the baselines.}
\label{fig:p5_st}
\end{figure}
\end{center}

\begin{center}
\begin{figure}
\begin{subfigure}{.5\textwidth}
\centering
\includegraphics[width=\linewidth]{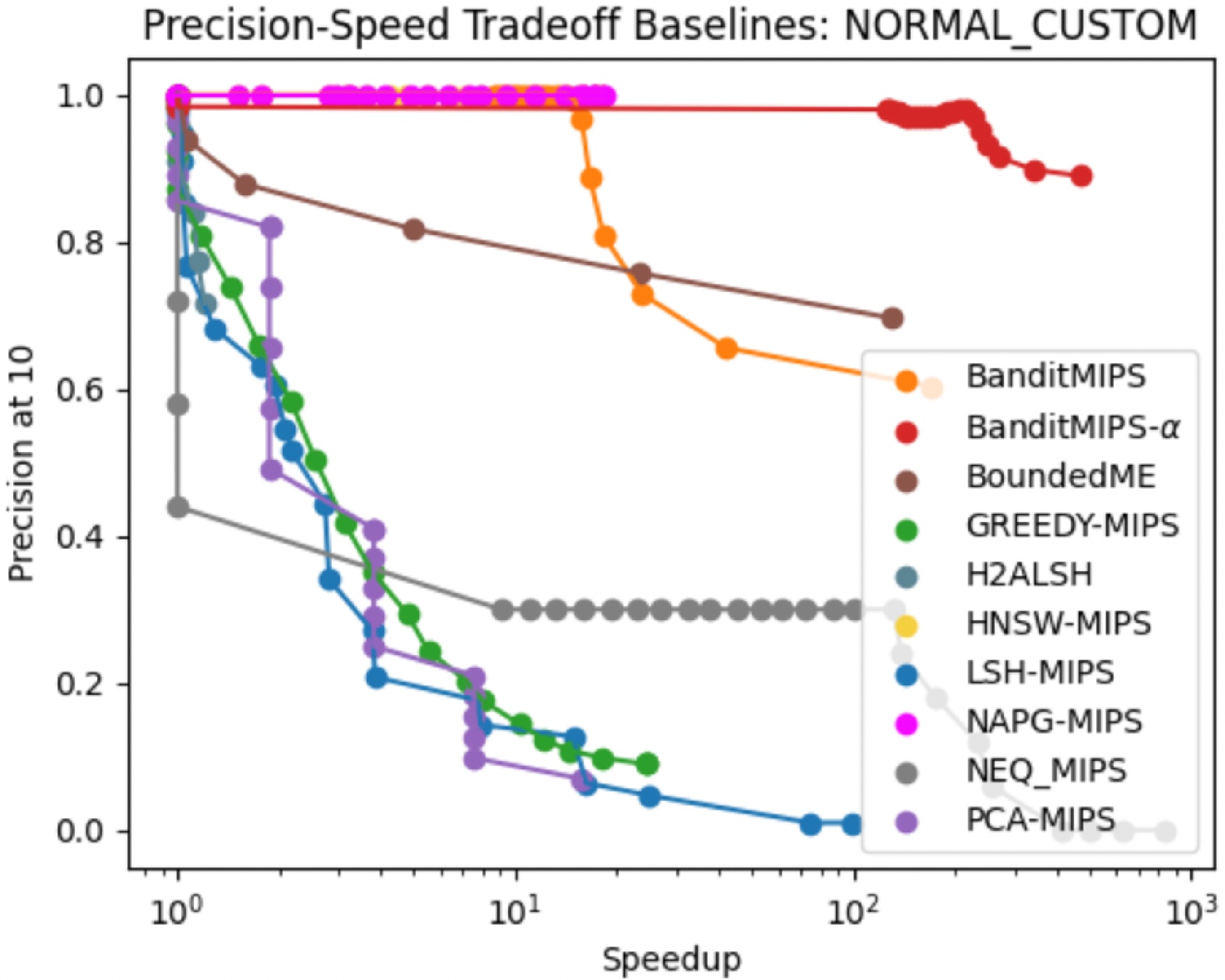}
\caption{}
\label{p10st:a}
\end{subfigure}
\begin{subfigure}{.5\textwidth}
\centering
\includegraphics[width=\linewidth]{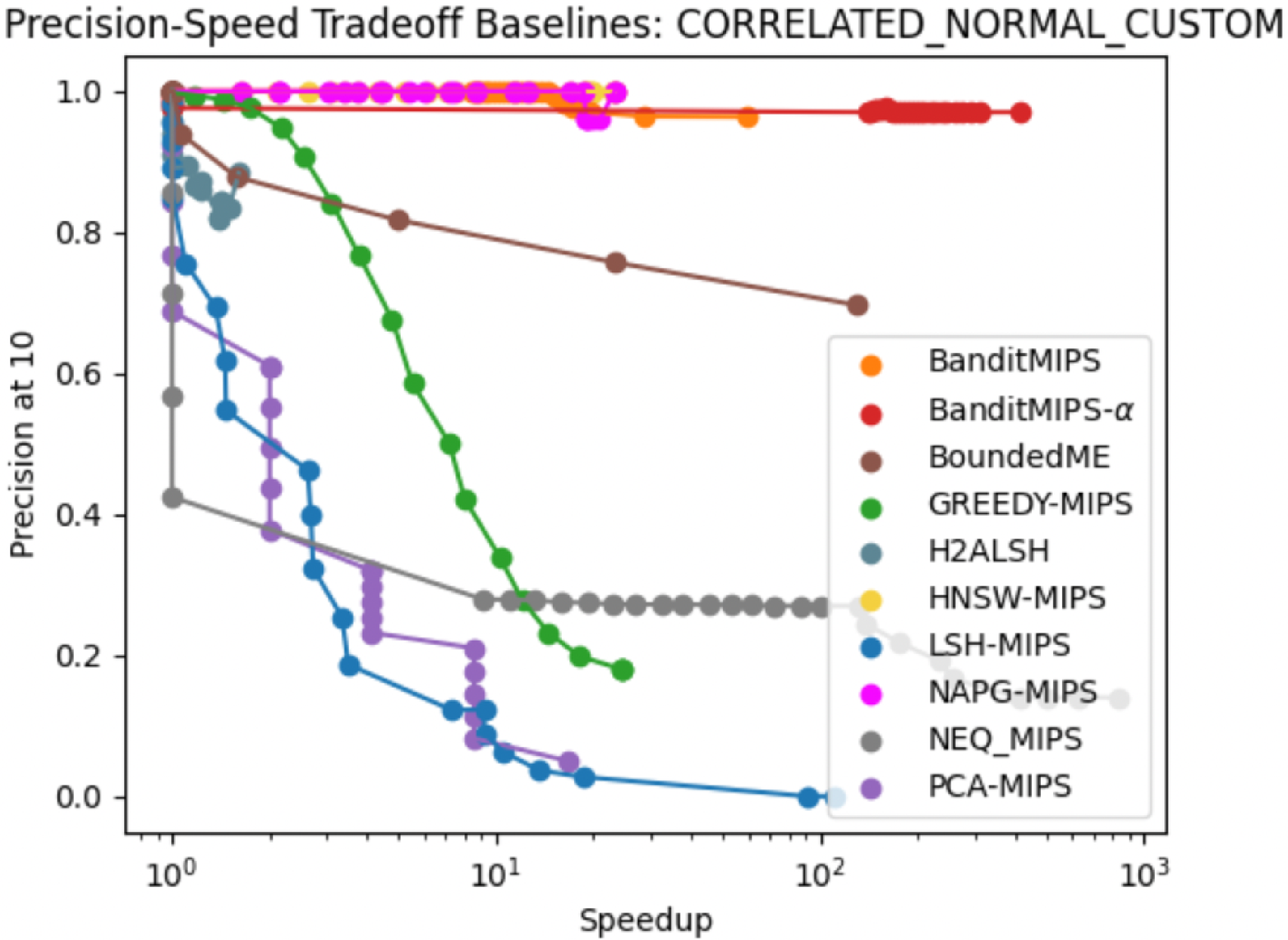}
\caption{}
\label{p10st:b}
\end{subfigure}
\begin{subfigure}{.5\textwidth}
\centering
\includegraphics[width=\linewidth]{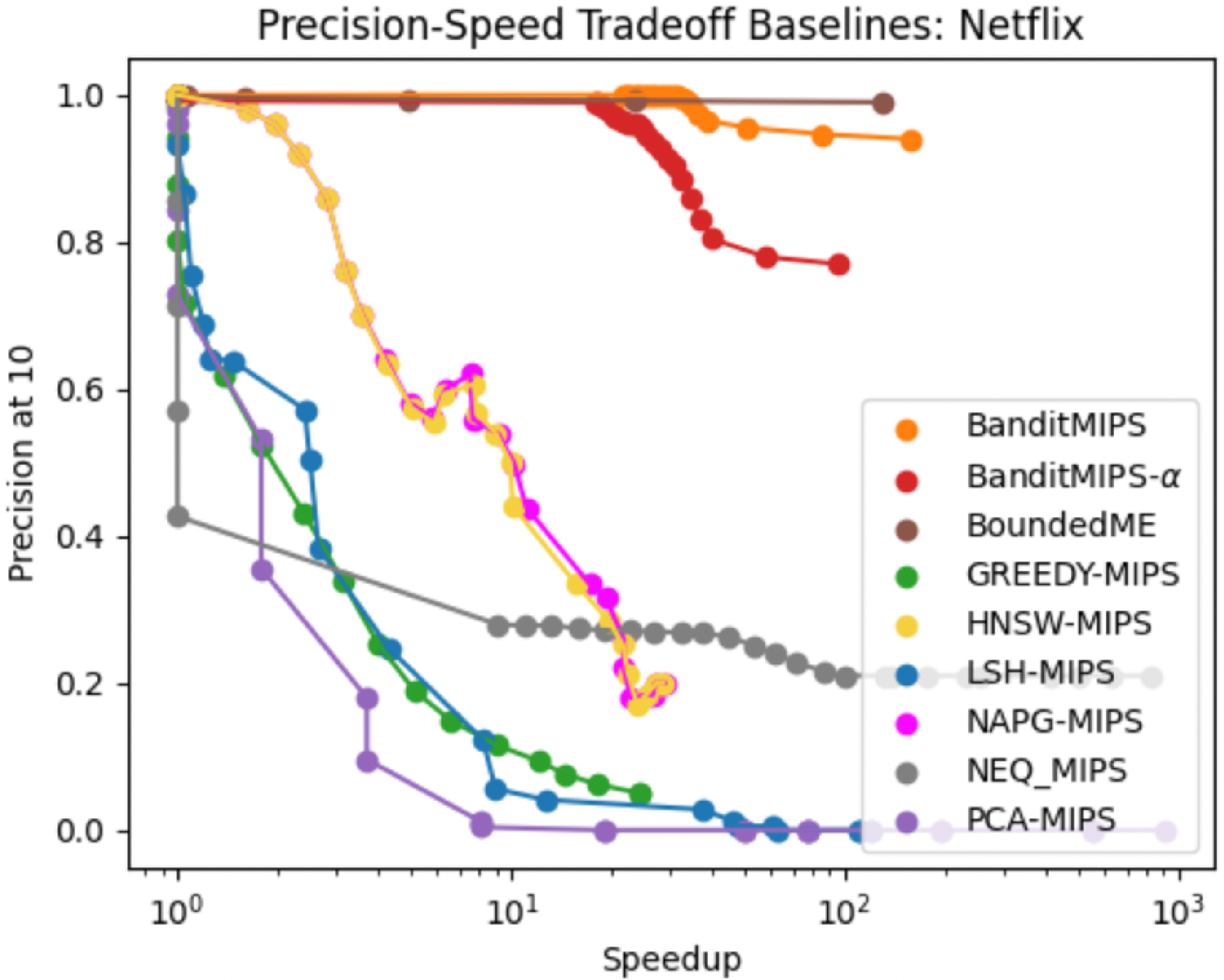}
\caption{}
\label{p10st:c}
\end{subfigure}
\begin{subfigure}{.5\textwidth}
\centering
\includegraphics[width=\linewidth]{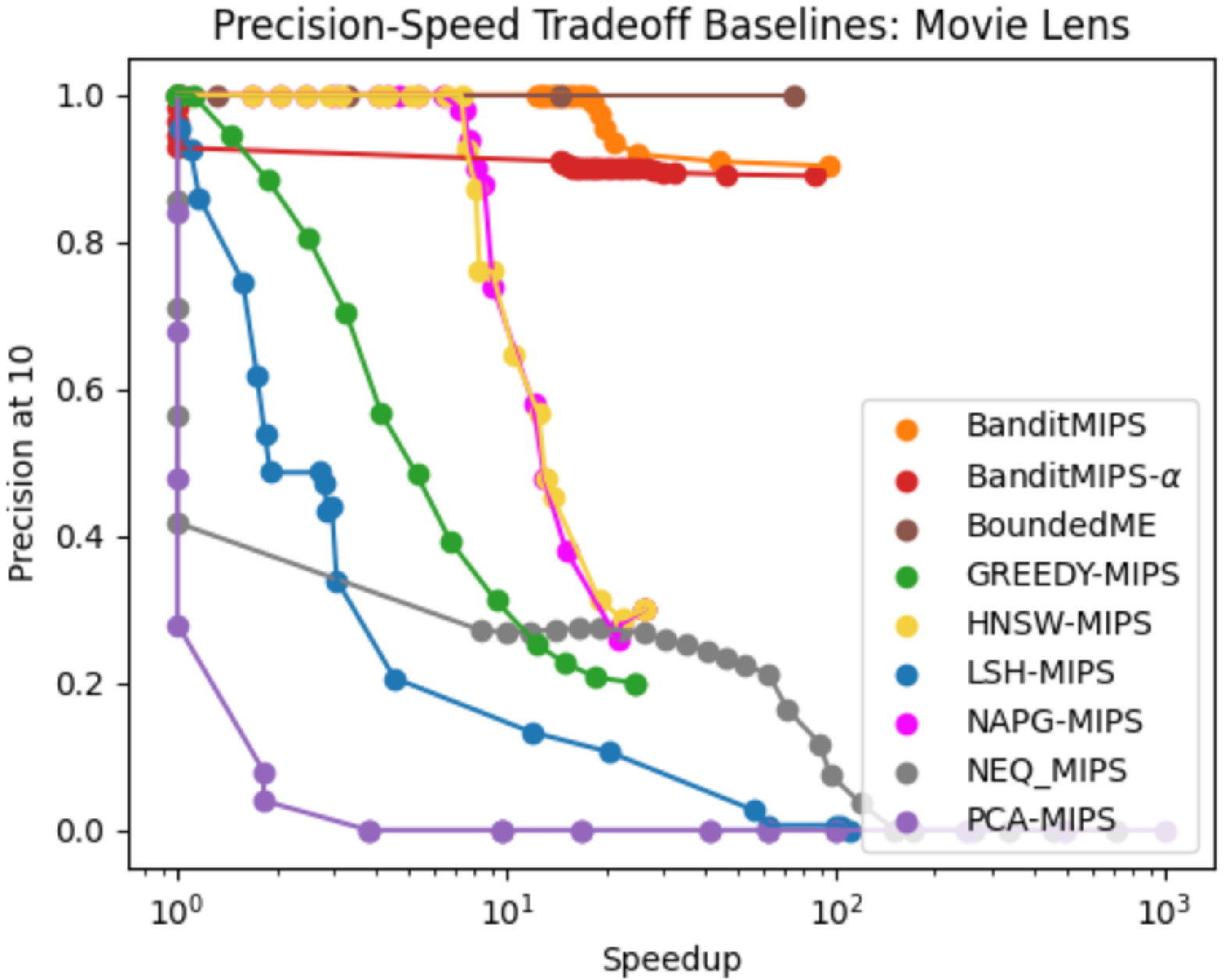}
\caption{}
\label{p10st:d}
\end{subfigure}
\caption{Trade-off between top-$10$ accuracy (precision@$10$) and speed for various algorithms across all four datasets. The $x$-axis represents the speedup relative to the naive $O(nd)$ algorithm and the $y$-axis shows the proportion of times an algorithm returned correct answer; higher is better. Each dot represents the mean across 10 random trials and the CIs are omitted for clarity. Our algorithms consistently achieve better accuracies at higher speedup values than the baselines.}
\label{fig:p10_st}
\end{figure}
\end{center}

Here, we present the results for precision@$k$ versus algorithm speedup for various algorithms for $k =5$ and $10$; see Figures \ref{fig:p5_st} and \ref{fig:p10_st}.
The precision@$k$ is defined as the proportion of true top $k$ atoms surfaced in the algorithm's returned top $k$ atoms (the precision@$k$ is also the top-$k$ accuracy, i.e., the proportion of correctly identified top $k$ atoms).
% \begin{figure}
% \centering
% \subfigure{\label{p5st:a} \includegraphics[width=.45\linewidth]{figures/p5st_nc.eps}}
% \subfigure{\label{p5st:b} \includegraphics[width=.45\linewidth]{figures/p5st_cnc.eps}}
% \subfigure{\label{p5st:c} \includegraphics[width=.45\linewidth]{figures/p5st_nf.eps}}
% \subfigure{\label{p5st:d} \includegraphics[width=.45\linewidth]{figures/p5st_ml.eps}} 
% \caption{Trade-off between top-$5$ accuracy (precision@$5$) and speed for various algorithms across all four datasets. The $x$-axis represents the speedup relative to the naive $O(nd)$ algorithm and the $y$-axis shows the proportion of times an algorithm returned correct answer; higher is better. Each dot represents the mean across 10 random trials and the CIs are omitted for clarity. Our algorithms consistently achieve better accuracies at higher speedup values than the baselines. 
% }
% \label{fig:p5_st}
% \end{figure}

% \begin{figure}
% \centering
% \subfigure{\label{p10st:a} \includegraphics[width=.45\linewidth]{figures/p10st_nc.eps}}
% \subfigure{\label{p10st:b} \includegraphics[width=.45\linewidth]{figures/p10st_cnc.eps}}
% \subfigure{\label{p10st:c} \includegraphics[width=.45\linewidth]{figures/p10st_nf.eps}}
% \subfigure{\label{p10st:d} \includegraphics[width=.45\linewidth]{figures/p10st_ml.eps}} 
% \caption{Trade-off between top-$10$ accuracy (precision@$10$) and speed for various algorithms across all four datasets. The $x$-axis represents the speedup relative to the naive $O(nd)$ algorithm and the $y$-axis shows the proportion of times an algorithm returned correct answer; higher is better. Each dot represents the mean across 10 random trials and the CIs are omitted for clarity. Our algorithms consistently achieve better accuracies at higher speedup values than the baselines. 
% }
% \label{fig:p10_st}
% \end{figure}

%% file: 9.5-appendix_preprocessing.tex
% !TEX root = 0-main.tex

\section{Using Preprocessing Techniques with \algnamenospace}
\label{app:preprocessing}

In this Appendix, we discuss using preprocessing techniques with \algnamenospace. The specific form of preprocessing is binning by estimated norm. More precisely, we estimate the norm of each atom by sampling a constant number of coordinates from them. We then sort the atoms by estimated norm into bins, where each bin contains $k = 30$ atoms (note that $k$ is a hyperparameter). The top $k$ atoms with the highest estimated norm are sorted into the first bin, the next $k$ atoms are sorted into the second bin, and so on. 

When running \algnamenospace, we find the best atom in each bin but stop sampling an atom if the best atom we have found across all bins has a sampled inner product greater than another atom's \textit{maximum} potential inner product. Intuitively, this allows us to filter atoms with small estimated norm more quickly if an atom in another bin is very likely to be a better candidate.

\begin{center}
\begin{figure}[ht]
\begin{subfigure}{.5\textwidth}
\centering
\includegraphics[width=\linewidth]{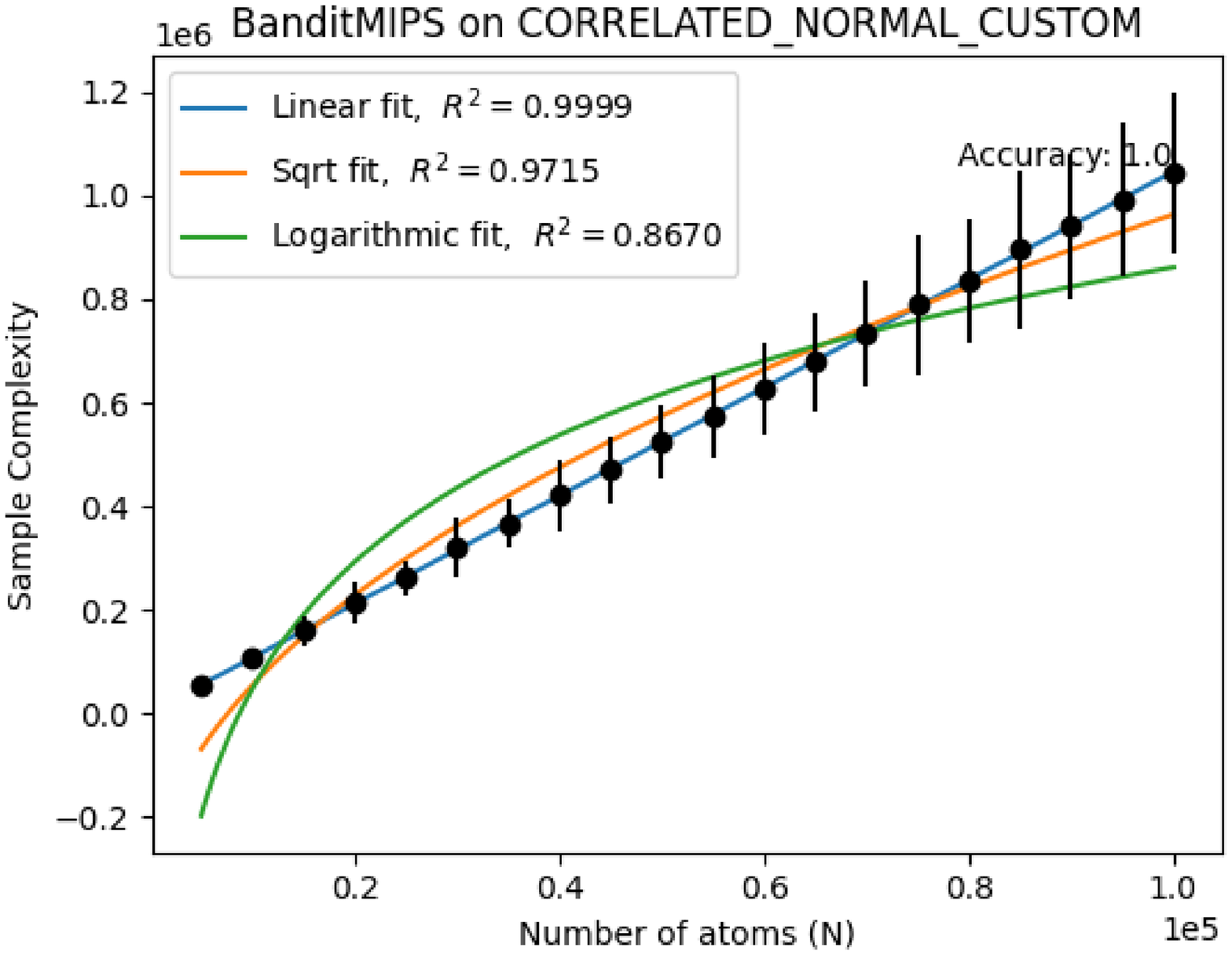}
\caption{}
\label{fig:bm_scaling_n_cnc}
\end{subfigure}
\begin{subfigure}{.5\textwidth}
\centering
\includegraphics[width=\linewidth]{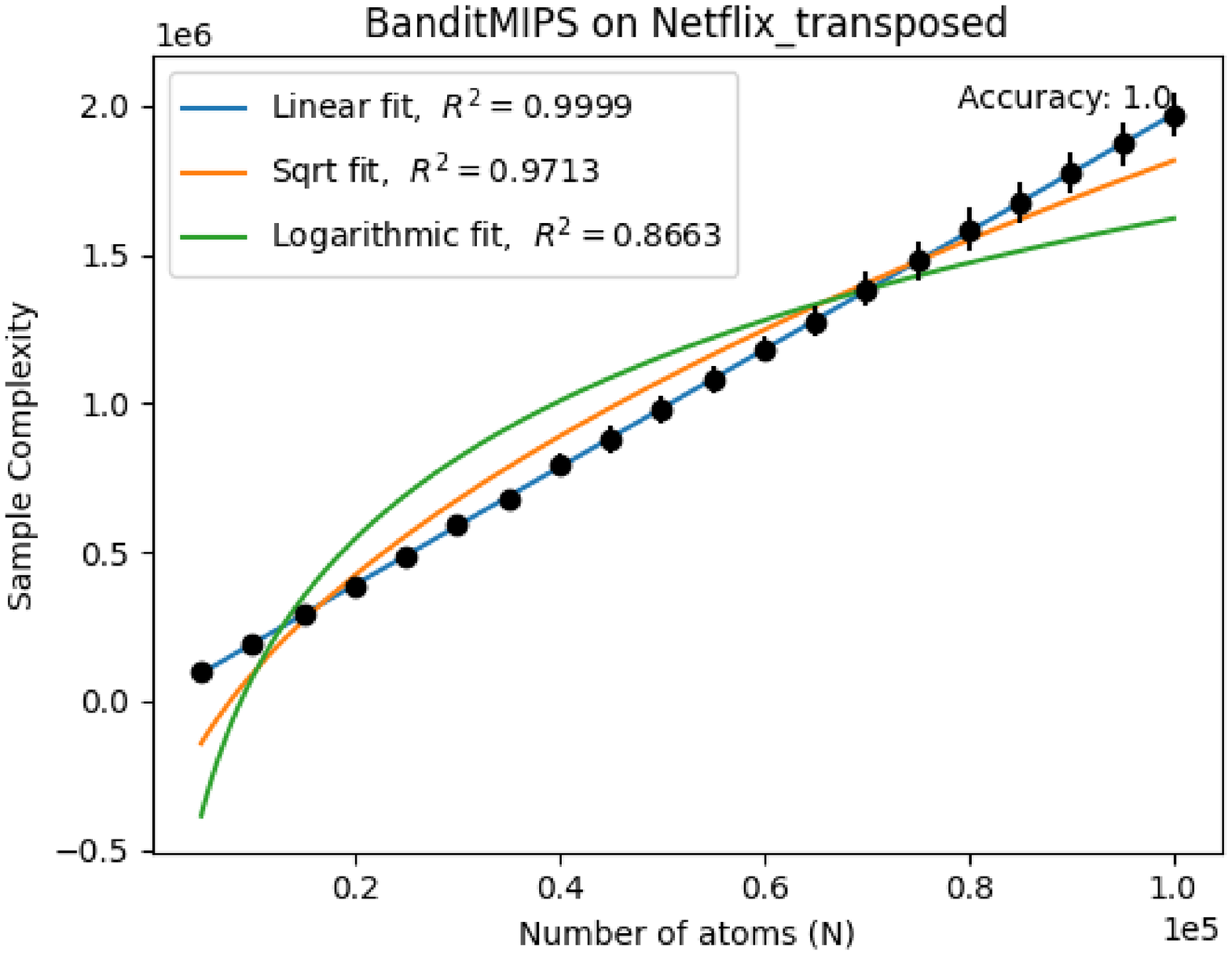}
\caption{}
\label{fig:bm_scaling_n_nf}
\end{subfigure}
\begin{subfigure}{.5\textwidth}
\centering
\includegraphics[width=\linewidth]{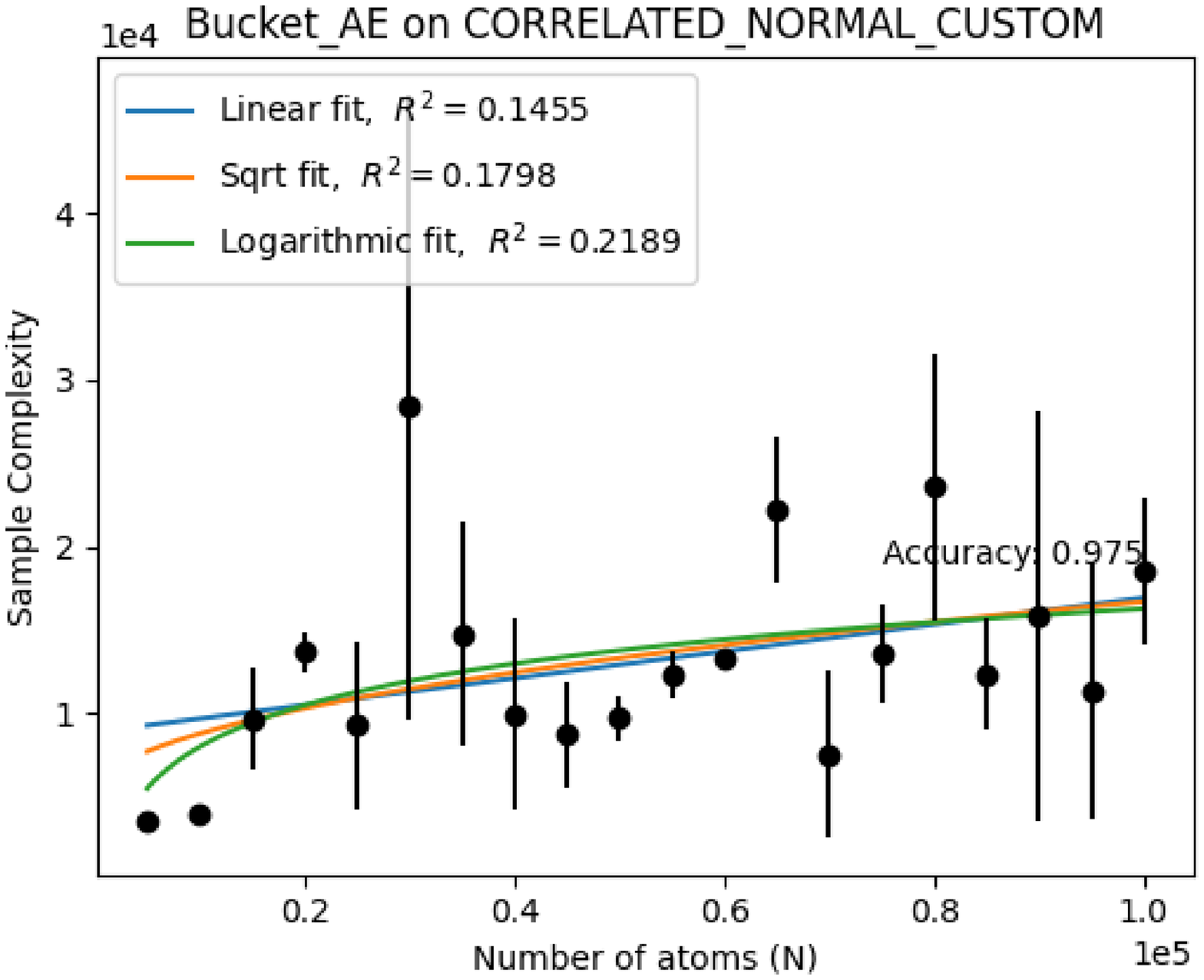}
\caption{}
\label{fig:bucket_scaling_n_cnc}
\end{subfigure}
\begin{subfigure}{.5\textwidth}
\centering
\includegraphics[width=\linewidth]{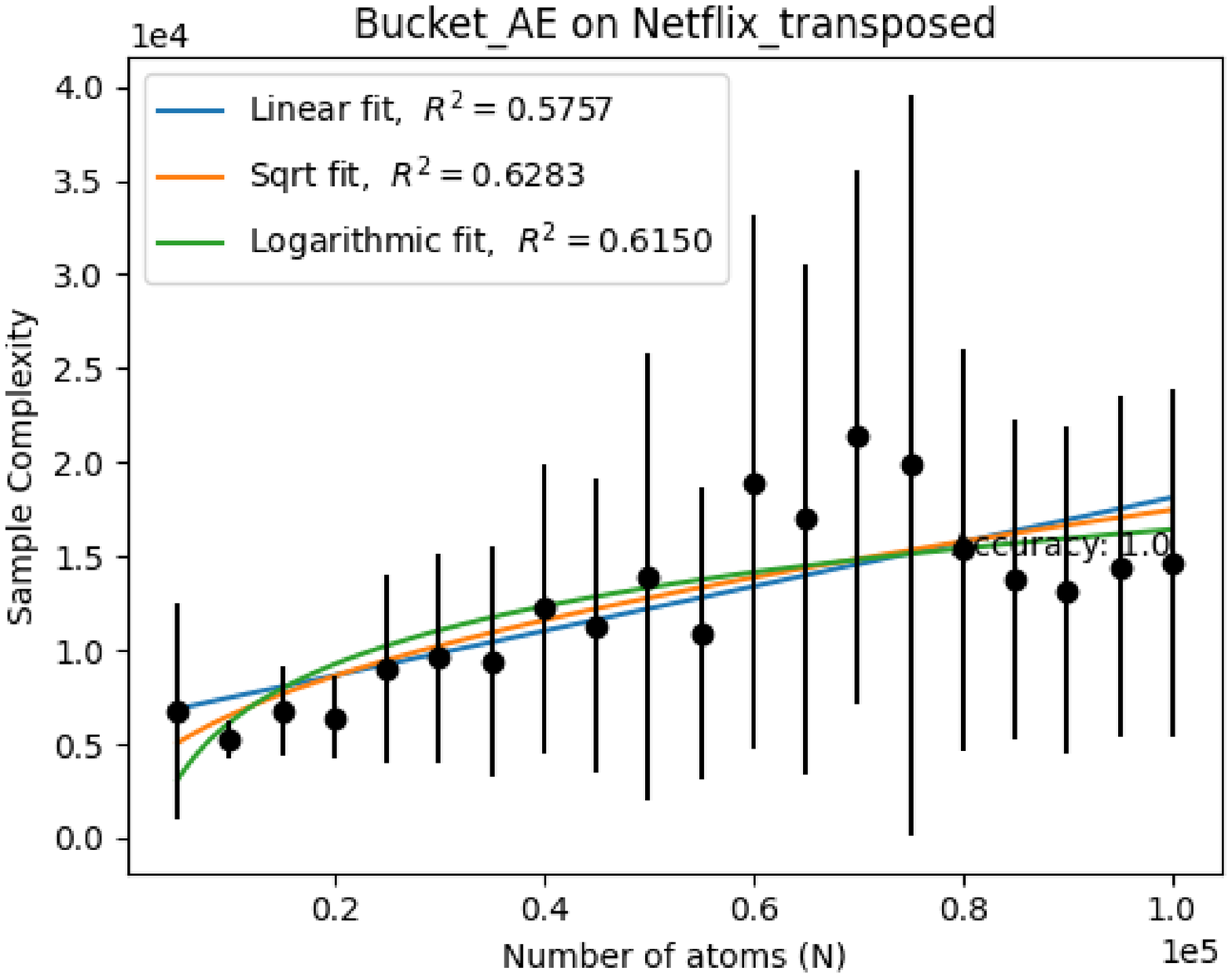}
\caption{}
\label{fig:bucket_scaling_n_nf}
\end{subfigure}
\begin{subfigure}{.5\textwidth}
\centering
\includegraphics[width=\linewidth]{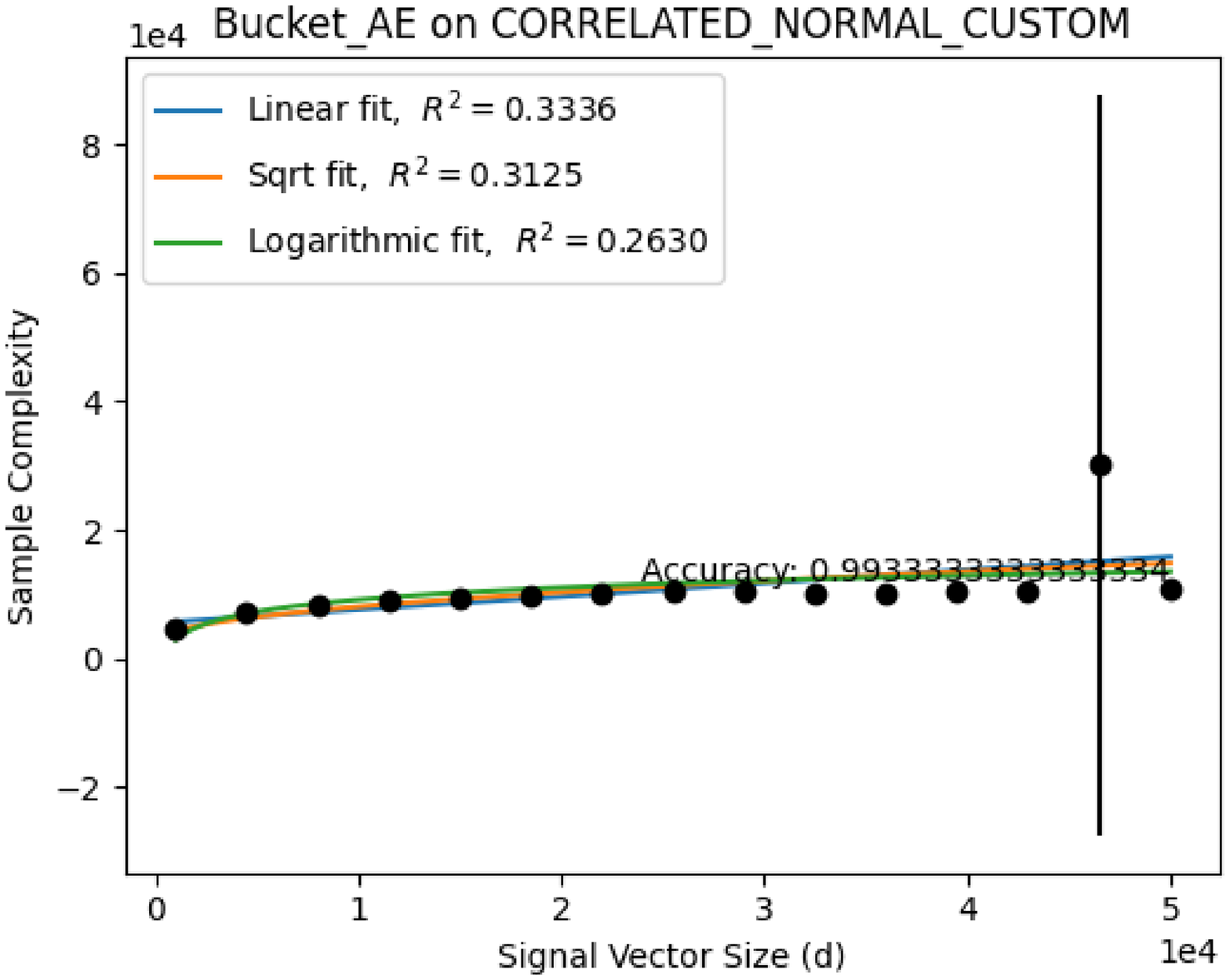}
\caption{}
\label{fig:bucket_scaling_d_cnc}
\end{subfigure}
\begin{subfigure}{.5\textwidth}
\centering
\includegraphics[width=\linewidth]{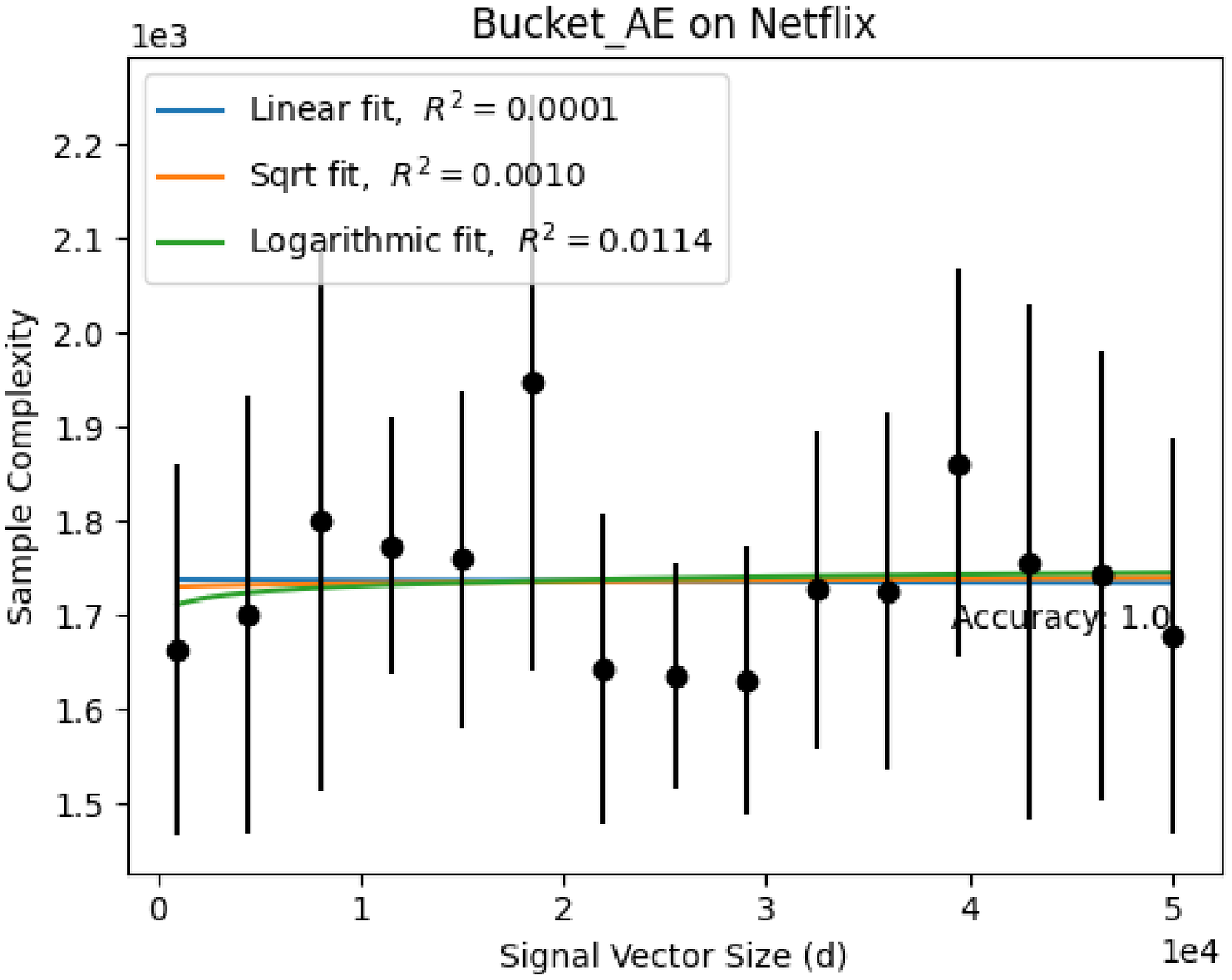}
\caption{}
\label{fig:bucket_scaling_d_nf}
\end{subfigure}
\caption{Top left and top right: sample complexity of \algname versus $n$ for the CORRELATED\_NORMAL\_CUSTOM and Netflix datasets. \algname scales linearly with $n$. Middle left and middle right: \algname with preprocessing Bucket\_AE scales sublinearly with $n$. This suggests that the form of preprocessing we apply is useful for reducing the complexity of our algorithm with $n$. Bottom left and bottom right: Bucket\_AE still scales as $O(1)$ with $d$. Means and uncertainties were obtained from 10 random seeds.}
\label{fig:preprocessing}
\end{figure}
\end{center}

We call this algorithm (\algname with this form of preprocessing) \texttt{Bucket\_AE}. Figure \ref{fig:preprocessing} demonstrates that \texttt{Bucket\_AE} reduces the scaling with $n$ of \algnamenospace. Furthermore, \texttt{Bucket\_AE} still scales as $O(1)$ with $d$.

We leave an exact complexity analysis of this preprocessing's affect on the scaling with $n$ under various distributional assumptions to future work.

% \begin{figure}
% \centering
% \subfigure{\label{fig:bm_scaling_n_cnc} \includegraphics[width=.45\linewidth]{figures/bm_scaling_n_cnc.eps}}
% \subfigure{\label{fig:bm_scaling_n_nf} \includegraphics[width=.45\linewidth]{figures/bm_scaling_n_nf.eps}}
% \subfigure{\label{fig:bucket_scaling_n_cnc} \includegraphics[width=.45\linewidth]{figures/bucket_scaling_n_cnc.eps}}
% \subfigure{\label{fig:bucket_scaling_n_nf} \includegraphics[width=.45\linewidth]{figures/bucket_scaling_n_nf.eps}} 
% \subfigure{\label{fig:bucket_scaling_d_cnc} \includegraphics[width=.45\linewidth]{figures/bucket_scaling_d_cnc.eps}}
% \subfigure{\label{fig:bucket_scaling_d_nf} \includegraphics[width=.45\linewidth]{figures/bucket_scaling_d_nf.eps}} 
% \caption{Top left and top right: sample complexity of \algname versus $n$ for the \texttt{CORRELATED\_NORMAL\_CUSTOM} and Netflix datasets. \algname scales linearly with $n$. Middle left and middle right: \algname with preprocessing \texttt{Bucket\_AE} scales sublinearly with $n$. This suggests that the form of preprocessing we apply is useful for reducing the complexity of our algorithm with $n$. Bottom left and bottom right: \texttt{Bucket\_AE} still scales as $O(1)$ with $d$. Means and uncertainties were obtained from 10 random seeds.}
% \label{fig:preprocessing}
% \end{figure}

%% file: 9.6-appendix_high_dim.tex
% !TEX root = 0-main.tex

\section{Experiments High Dimensional Datasets and Application to Matching Pursuit}
\label{app:high_dim}

One of the advantages of \algname is that it has no dependence on dataset dimensionality when the necessary assumptions are satisfied. 
%In this section, we describe the high-dimensional real-world datasets used in Section \ref{sec:experiments}.
We also demonstrate the $O(1)$ scaling with $d$ of \algname explicitly on a high-dimensional synthetic dataset and discuss an application to the Matching Pursuit problem (MP).

\subsection{Application to Matching Pursuit in High Dimensions: the \texttt{SimpleSong} Dataset}

\subsubsection{Description of the \texttt{SimpleSong} Dataset}
We construct a simple synthetic dataset, entitled the \texttt{SimpleSong} Dataset. In this dataset, the query and atoms are audio signals sampled at 44,100 Hz and each coordinate value represents the signal's amplitude at a given point in time. Common musical notes are represented as periodic sine waves with the frequencies given in Table \ref{table:notes}.

\begin{table}
\caption{Frequencies for various musical notes.} 
\vspace{1em}
\label{table:notes}
\centering
\begin{tabular}{ll}
\toprule
Note & Frequency (Hz) \\
\midrule
C4   & 256            \\
E4   & 330            \\
G4   & 392            \\
C5   & 512            \\
E5   & 660            \\
G5   & 784            \\
\bottomrule
\end{tabular}
\end{table}

The query in this dataset is a simple song. The song is structured in 1 minute intervals, where the first interval -- called an A interval --  consists of a C4-E4-G4 chord and the second interval -- called a B interval --  consists of a G4-C5-E5 chord. The song is then repeated $t$ times, bringing its total length to $2t$ minutes. The dimensionality of the the signal is $d = 2t * 44,100 = 88,200t$. The weights of the C4, E4, and G4 waves in the A intervals and the G4, C5, and E5 waves in the B intervals are in the ratio 1:2:3:3:2.5:1.5. 

The atoms in this dataset are the sine waves corresponding to the notes with the frequencies show in Table \ref{table:notes}, as well as notes of other frequencies. 

\subsubsection{Matching Pursuit and Fourier Transforms}

The Matching Pursuit problem (MP) is a problem in which a vector $\mathbf{q}$ is approximated as a linear combination of the atoms $\mathbf{v}_1, \ldots, \mathbf{v}_n$. A common algorithm for MP involves solving MIPS to find the atom $\mathbf{v}_{i^*}$ with the highest inner product with the query, subtracting the component of the query parallel to $\mathbf{v}_{i^*}$, and re-iterating this process with the residual. Such an approach solves MIPS several times as a subroutine. Thus, an algorithm which accelerates MIPS should also then accelerate MP. 

In the audio domain, we note that when the atoms $\mathbf{v}_1, \ldots, \mathbf{v}_n$ are periodic functions with predefined frequencies, MP becomes a form of Fourier analysis in which the atoms are the Fourier components and their inner products with the query correspond to Fourier coefficients. For more detailed background on Fourier theory, we refer the reader to \cite{oran1988}.

\subsubsection{Experimental Results}

For convenience, we restrict $t$ to be an integer in our experiments so a whole number of AB intervals are completed. We ran \algname with $\delta = \frac{1}{10,000}$ and $\sigma^2 = 6.25$ over $3$ random seeds for various values of $t$.

\begin{figure}
    \centering
    \includegraphics[scale=0.7]{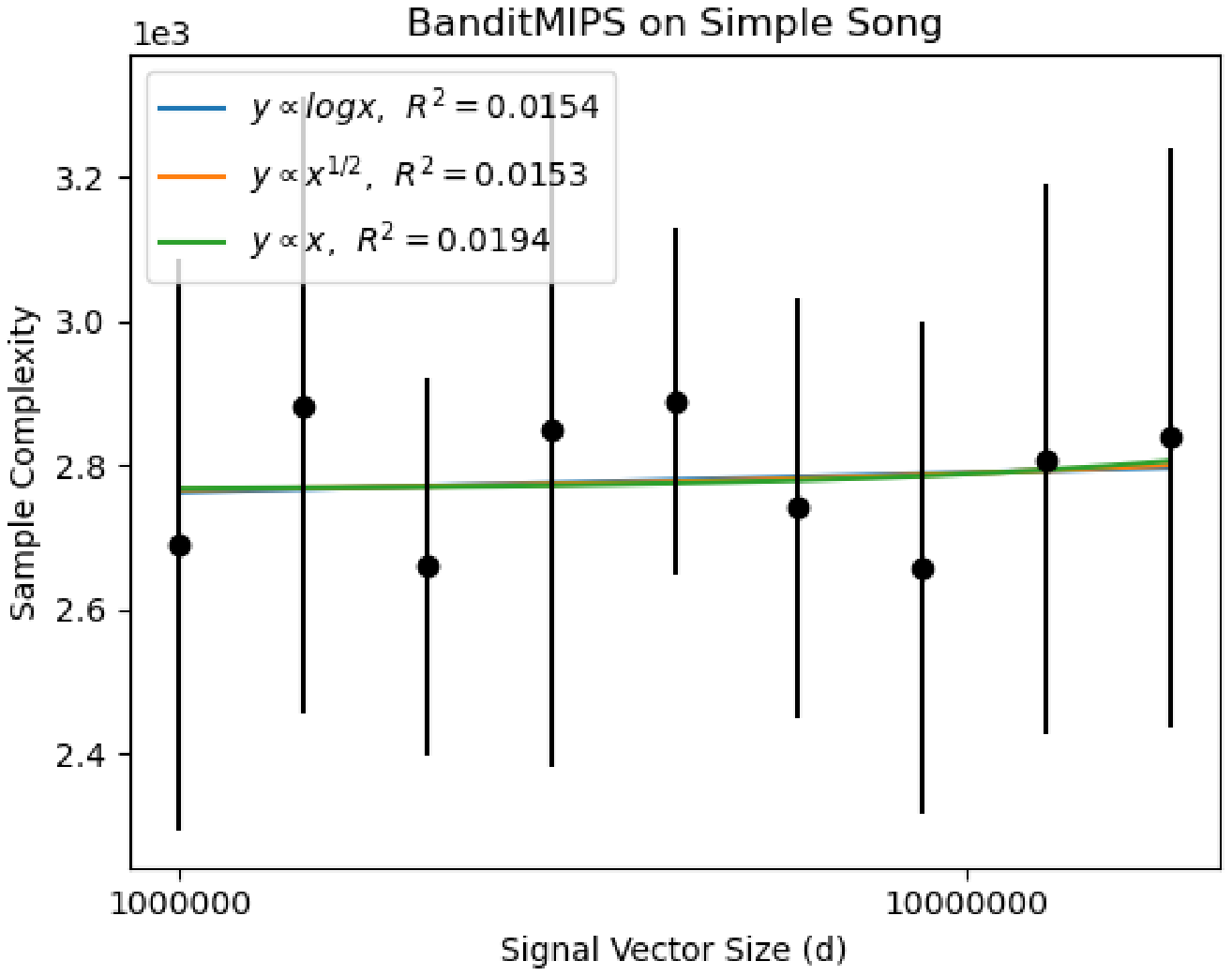}
    \caption{Sample complexity of MP when using \algname as a subroutine for MIPS on the \texttt{SimpleSong} dataset. The complexity of the solving the problem does not scale with the length of the song, $d$. Uncertainties and means were obtained from 3 random seeds. \algname returns the correct solution to MIPS in each trial.}
    \label{fig:simple_song_scaling}
\end{figure}

\algname is correctly able to recover the notes played in the song in order of decreasing strength: G4, C5, E4, E5, and C4 in each experiment. 
Furthermore, \algname is able to calculate their Fourier coefficients correctly.
Crucially, the complexity of \algname to identify these components does not scale with $d$, the length of the song.
%Intuitively, this example is similar to the one in Figure \ref{fig:toy_problem}.
Figure \ref{fig:simple_song_scaling} demonstrates the total sample complexity of \algname to identify the first five Fourier components (five iterations of MIPS) of the song as the song length increases.

Our approach may suggest an application to Fourier transforms, which aim to represent signals in terms of constituent signals with predetermined set of frequencies. We acknowledge, however, that Fourier analysis is a well-developed field and that further research is necessary to compare such a method to state-of-the-art Fourier transform methods, which may already be heavily optimized or sampling-based.

%% file: 9.7-appendix_symmetric.tex
% !TEX root = 0-main.tex

\section{\algname on a Highly Symmetric Dataset}
\label{app:symmetric}

In this section, we discuss a dataset on which the assumptions in Section \ref{subsec:gaps} fail, namely when $\Delta$ scales with $d$. In this setting, \algname does not scale as $O(1)$ and instead scales linearly with $d$, as is expected.

We call this dataset the \texttt{SymmetricNormal} dataset. In this dataset, the signal has each coordinate drawn from $\mathcal{N}(0, 1)$ and each atom's coordinate is drawn i.i.d. from $\mathcal{N}(0, 1)$. Note that all atoms are therefore symmetric \textit{a priori}.

We now consider the quantity $\Delta_{i,j}(d) \coloneqq \mu_1(d) - \mu_2(d)$, i.e., the gap between the first and second arm, where our notation emphasizes we are studying each quantity as $d$ increases. 
Note that $\Delta_{i,j}(d) = \frac{\mathbf{v}_1^T q - \mathbf{v}_2^T q}{d}$. 
By the Central Limit Theorem, the sequence of random variables $\sqrt{d}\Delta_{i,j}(d)$ converges in distribution to $\mathcal{N}(0, \sigma^2_{i,j})$ for some constant $\sigma^2_{i,j}$.
Crucially, this implies that $\Delta_{i,j}(d)$ is on the order of $\frac{1}{\sqrt{d}}$. 

\begin{figure}
    \centering
    \includegraphics[scale=0.7]{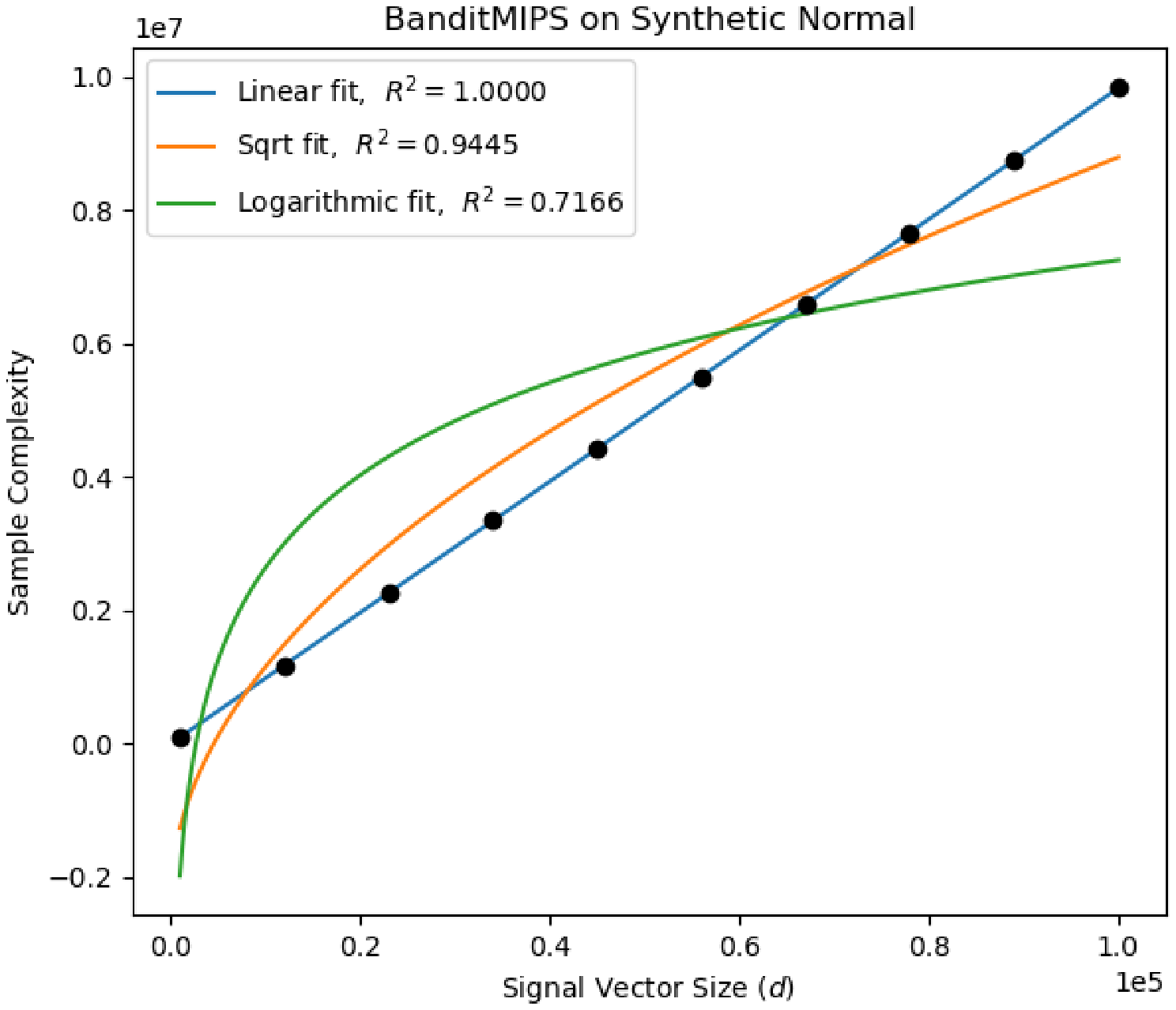}
    \caption{Sample complexity of \algname on the \texttt{SymmetricNormal} dataset. The sample complexity of \algname is linear with $d$, as is expected. Uncertainties and means were obtained from 10 random seeds.}
    \label{fig:bm_sn}
\end{figure}

The complexity result from Theorem \ref{thm:specific} then predicts that \algname scales linearly with $d$. Indeed, this is what we observe in Figure \ref{fig:bm_sn}.

